\documentclass[runningheads]{llncs}

\usepackage{eccv}

\usepackage{eccvabbrv}

\usepackage{graphicx}
\usepackage{booktabs}

\usepackage[accsupp]{axessibility}  

\usepackage{hyperref}

\usepackage{orcidlink}
\graphicspath{{./figures/}{./supp_figs/}{./rebt_figs/}}

\usepackage[group-separator={,}]{siunitx}
\usepackage{amsmath}
\usepackage{amsfonts}
\usepackage{makecell}
\usepackage{lipsum}
\usepackage{tcolorbox}
\usepackage{multirow}
\usepackage{marvosym}

\usepackage[capitalize]{cleveref}
\crefname{section}{Sec.}{Secs.}
\Crefname{section}{Section}{Sections}
\Crefname{table}{Table}{Tables}
\crefname{table}{Tab.}{Tabs.}
\Crefname{figure}{Figure}{Figures}
\crefname{figure}{Fig.}{Figs.}
\Crefname{equation}{Equation}{Equations}
\crefname{equation}{Eq.}{Eqs.}
\Crefname{algorithm}{Algorithm}{Algorithms}
\crefname{algorithm}{Alg.}{Algs.}

\usepackage[labelformat=simple]{subcaption}

\colorlet{lightpink}{pink!35}
\colorlet{lightcyan}{cyan!20}
\colorlet{lightgray}{gray!60}
\definecolor{darkgray}{rgb}{0.8, 0.8, 0.8}
\colorlet{red}{red!80}
\colorlet{blue}{blue!80}
\colorlet{green}{green!60!black}
\colorlet{algemp}{cyan!10}

\usepackage{dirtree}
\usepackage{xcolor,colortbl}
\usepackage{pifont}
\usepackage{wrapfig}

\newcolumntype{a}{>{\columncolor{gray!20!white}}c}

\newcommand{\xres}{X_\textup{res}}
\newcommand{\xattn}{X_\textup{attn}}
\newcommand{\xsum}{X_\textup{sum}}
\newcommand{\id}{\mathbb{I}}

\begin{document}

\title{ClearCLIP: Decomposing CLIP Representations \\ for Dense Vision-Language Inference} 

\titlerunning{ClearCLIP}

\author{Mengcheng Lan\inst{1} \and
Chaofeng Chen\inst{1} \and
Yiping Ke\inst{2} \and 
Xinjiang Wang \inst{3} \and \\
Litong Feng \inst{3}\thanks{Corresponding author.} \and
Wayne Zhang \inst{3}}

\authorrunning{M.Lan et al.}

\institute{S-Lab, Nanyang Technological University \and
CCDS, Nanyang Technological University\ \ \ \ 
\inst{3} SenseTime Research \\
\email{lanm0002@e.ntu.edu.sg}\ \  
\email{\{chaofeng.chen,\ ypke\}@ntu.edu.sg}\\
\email{\{wangxinjiang, fenglitong, wayne.zhang\}@sensetime.com}
\url{https://github.com/mc-lan/ClearCLIP}
}

\maketitle

\begin{abstract}
    Despite the success of large-scale pretrained Vision-Language Models (VLMs) especially CLIP in various open-vocabulary tasks, their application to semantic segmentation remains challenging, producing noisy segmentation maps with mis-segmented regions. 
    In this paper, we carefully re-investigate the architecture of CLIP, and identify residual connections as the primary source of noise that degrades segmentation quality.
    With a comparative analysis of statistical properties in the residual connection and the attention output across different pretrained models, we discover that CLIP's image-text contrastive training paradigm emphasizes global features at the expense of local discriminability, leading to noisy segmentation results. 
    In response, we propose ClearCLIP, a novel approach that decomposes CLIP's representations to enhance open-vocabulary semantic segmentation. 
    We introduce three simple modifications to the final layer: removing the residual connection, implementing the self-self attention, and discarding the feed-forward network. 
    ClearCLIP consistently generates clearer and more accurate segmentation maps and outperforms existing approaches across multiple benchmarks, affirming the significance of our discoveries.
\end{abstract}

\section{Introduction}
\label{sec:intro}
\begin{figure}[t]
    \centering
    \includegraphics[width=0.9\linewidth]{ 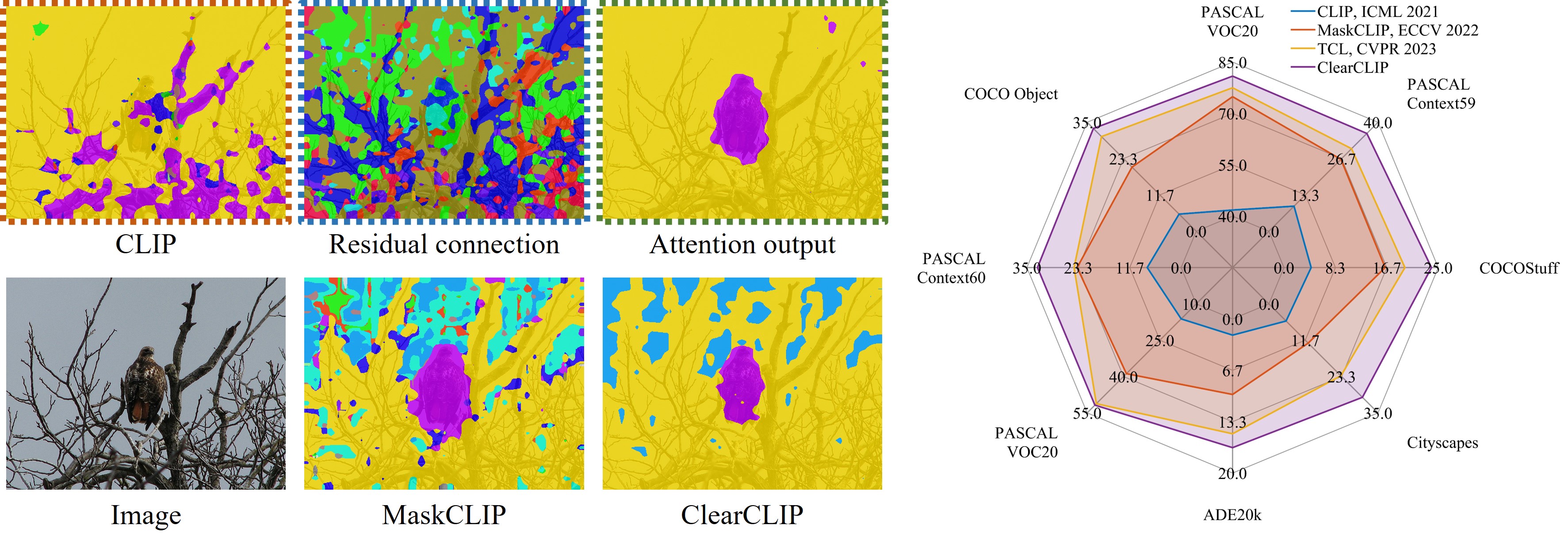}
    \caption{Left: Example of open-vocabulary semantic segmentation. CLIP \cite{radford2021learning} fails to localize the object. MaskCLIP \cite{zhou2022extract} can localize the foreground and background but still exhibits significant noise. Our proposed method, ClearCLIP, achieves high-quality segmentation map. Our key insight is that vanilla CLIP's segmentation map can be decomposed into a cluttered map of residual connection and a clearer and smoother map of attention output from the last transformer layer. Right: Comparison of open-vocabulary semantic segmentation performance. 
    }
    \label{fig:teaser}
\end{figure}

Large-scale Vision-Language pre-trained Models (VLMs), represented by the Contrastive Language-Image Pre-training (CLIP) family \cite{radford2021learning, cherti2023reproducible, xu2023demystifying}, have demonstrated remarkable generality and robustness across a diverse range of downstream tasks, \eg, zero-shot image classification \cite{jia2021scaling, radford2021learning}, visual question answering \cite{antol2015vqa, yu2022coca, khan2022weakly} and image-text retrieval \cite{mishra2013image, cho2021unifying, li2022blip}. 
There is a growing interest in leveraging the power of CLIP for open-vocabulary and zero-shot problems.
However, CLIP falls behind to maintain its zero-shot capabilities for dense prediction tasks especially semantic segmentation, as depicted by previous works \cite{zhou2022extract,li2023clip}. 
This limitation arises primarily from the training data used for VLMs, which consists largely of image-level labels and lacks sensitivity to visual localization.
For example, as shown in \cref{fig:teaser} (top-left image), the segmentation map generated by CLIP based on patch-level cosine similarity between visual and textual features reveals many misclassified patches and significant noise, showcasing the limitation of CLIP in dense visual localization. 
Recent works \cite{zhou2022extract,li2023clip, wang2023sclip,bousselham2023grounding} usually attribute noisy representations in CLIP to the self-attention layers, and have achieved great progress by improving this module. 
MaskCLIP \cite{zhou2022extract} adopted a simple technique of setting  \textit{query-key} attention map of the last block as an identical matrix, resulting in an improvement in mIoU on COCOStuff \cite{caesar2018coco} from 4.4 to 16.4.
CLIPSurgery \cite{li2023clip} argued that the \textit{value-value} attention map is cleaner and further improved the mIoU to 21.9. 
The recent study SCLIP \cite{wang2023sclip} combined the \textit{query-query} and \textit{value-value} attention and achieved even better results. 
While these enhancements in attention mechanisms lead to prioritization of relevant context and improved performance, their segmentation results still exhibit some noise. 
Such noise becomes more pronounced and leads to deteriorated performance when larger backbones are applied.
These raise the fundamental question: \textit{where do these noises originate and how do they surface in the CLIP models}?

To answer the \textit{where} question, we conduct a thorough investigation into the architecture of CLIP.
We are surprised to find that the residual connection, proposed by ResNet and commonly employed in transformer architectures, has a significant effect on the adaptation of CLIP to open-vocabulary semantic segmentation.
To elucidate this, we decompose the output of CLIP's vision encoder into two components: the residual connection and the attention output, which is achieved by directly separating them in the last layer of the vision encoder. 
As illustrated in \cref{fig:teaser} (top three images), the segmentation result obtained from the residual connection exhibits noticeable noise, while the attention output features produce significantly clearer results with superior localization properties. 
From these observations, we propose that \textit{the noises present in the segmentation map mainly originate from the residual connection.}

To delve into \textit{how} these noises emerge, we begin by comparing the statistical properties of CLIP's residual connection and attention output. 
Notably, we observe a significant discrepancy in their normalized entropy in CLIP: while the entropy of the residual connection tends to be near 0 along layers, the entropy of the attention output remains at 1. 
This finding aligns with our observation that its residual connection contains much larger maximum values along layers.
Accordingly, the final output of CLIP, \ie, the addition of residual connection and attention output, exhibits similar properties to the residual connection. 
Our findings also resonate with those of \cite{darcet2023vision}, which identify many artifacts with high-norm in the final feature maps of large-scale pretrained models. 
When we look closely at the residual connection maps, we see that these peak values are concentrated in a few channels. 
In other words, most feature vectors in the residual connection maps share the same peak dimensions, \ie, similar directions in the latent space. 
This property makes it difficult to distinguish each spatial feature vector through cosine similarity, 
thereby contributing to the generation of noises.
Conversely, the self-attention mechanism in the attention branch is learned to separate dissimilar spatial features, alleviating such issues. 
Next, we examine the DINO \cite{caron2021emerging}, which is also a transformer architecture but pretrained in a self-supervised way. 
We find that these two feature maps of DINO do not show such discrepancies in entropy.
Therefore, we propose that \textit{the high-level supervision in CLIP emphasizes the global feature direction in the residual latent space, making local feature vectors less distinguishable and leading to noise in residual features.}

Based on these discoveries, we revisit recent methods \cite{zhou2022extract, li2023clip, wang2023sclip} and find that the performance enhancements observed in these methods can be partially attributed to the reduced influence of the residual connection when the attention output is strengthened. 
We then deduce that two critical factors play a pivotal role in adapting CLIP for dense vision-language inference: the reduction in the impact of the residual connection and the reorganization of spatial information through the self-self attention. 
Guided by these insights, we introduce our approach, ClearCLIP, which incorporates three straightforward modifications to the final layer of CLIP: eliminating the residual connection, adopting the self-self attention, and discarding the Feed-Forward Network (FFN). These modifications are designed to boost the attention output, thereby producing a clearer representation for the task of open-vocabulary semantic segmentation, as shown in \cref{fig:teaser}. Extensive experiments on 8 benchmark datasets demonstrate the effectiveness of ClearCLIP.

\section{Related Work}
\label{sec:relatedwork}
\subsubsection{Vision-language pre-training.}
VLMs has experienced significant progress in recent years.
One notable family of vision-language models is based on contrastive learning \cite{alayrac2020self, jia2021scaling, li2021align, miech2020end, yao2021filip, yuan2021florence, radford2021learning, cherti2023reproducible, xu2023demystifying}.
Among them, CLIP \cite{radford2021learning} trained on a private WIT-400M with image-text
pairs achieves promising zero-shot capabilities for downstream tasks such as image-text retrieval, image classification via text prompts.
ALIGN \cite{jia2021scaling} adopts the same dual-encoder architecture as CLIP but is trained on a private dataset with over one billion noisy image-text pairs. 
Additionally, OpenCLIP \cite{cherti2023reproducible} explores scaling laws for CLIP by training the models on the public LAION \cite{schuhmann2022laion} dataset with up to two billion image-text pairs.
Another line of research \cite{kim2021vilt, li2022blip, chen2022pali} focuses on shared or mixed architectures between vision and language modalities, enabling additional zero-shot capabilities such as visual question answering \cite{kim2021vilt, khan2022weakly} and image captioning \cite{li2022blip}. 
Our work specifically addresses the adaptation of CLIP families \cite{radford2021learning,cherti2023reproducible}, for downstream dense prediction tasks.

\subsubsection{Open-vocabulary semantic segmentation.}
Open-vocabulary semantic segmentation, also known as zero-shot semantic segmentation, aims to segment an image with arbitrary categories described by texts.
Recent works have mainly built upon large-scale vision-language models \cite{radford2021learning, cherti2023reproducible, xu2023demystifying}, which could be roughly divided into three types.
1) \textbf{Training-free} methods \cite{zhou2022extract,li2023clip,bousselham2023grounding,li2024cascade,sun2024clip} attempt to tap into the inherent localization capabilities of CLIP with minimal modifications.
MaskCLIP \cite{zhou2022extract} proposes to extract the value embedding of the last self-attention bock of CLIP's vision encoder for dense prediction tasks.
Following this work, many studies \cite{li2023clip, bousselham2023grounding, wang2023sclip} generalize the query-key attention to a self-self attention mechanism, such as the value-value attention in CLIPSurgery \cite{li2023clip}, the query-query and key-key attention in SCLIP \cite{wang2023sclip}, and generalized self-self attention combination in GEM \cite{bousselham2023grounding}.
These modifications induce the model to focus more on relevant context, resulting in significantly improved performance.
2) \textbf{Unsupervised/Weakly-supervised} methods mainly involve the design of more intricate architectures aimed at explicitly grouping semantic contents with image-only/image-text training samples.
GroupViT \cite{xu2022groupvit} and SegCLIP \cite{luo2023segclip} introduce the grouping blocks into the vision encoder, whose group tokens serve as class centers for semantic segmentation.
OVSegmentor \cite{xu2023learning} also introduces a set of learnable group tokens via a slot-attention, and performs model training with masked entity completion and cross-image mask consistency proxy tasks.
Additionally, PGseg \cite{zhang2023uncovering} proposes to use both group tokens and prototype tokens to segment the images.
TCL \cite{cha2023learning} and CLIP-S$^4$ \cite{he2023clip} propose to directly generate mask/segment proposals within each image.
3) \textbf{Fully-supervised} methods usually involve in-domain fine-tuning, \eg, training on the COCOStuff\cite{caesar2018coco} training set with full dense annotations, and therefore typically achieve better performance compared to training-free and weakly-supervised methods.
Existing methods in this category can be broadly categorized into CLIP-based methods \cite{xu2022simple, jiao2023learning, yu2023convolutions, han2023open, xu2023masqclip, liang2023open} and Stable Diffusion-based methods \cite{li2023open, xu2023open, xu2023side}.

Our method belongs to training-free open-vocabulary semantic segmentation.
We aim to explore the intrinsic localization properties of CLIP from a perspective of feature decomposition.

\section{Methodology}
\label{sec:method}
In this section, we start by providing an overview of the CLIP model \cite{radford2021learning} and introducing a baseline for open-vocabulary dense inference in \cref{sec:preliminary}. Then, we show how the CLIP baseline fails to achieve satisfactory results which motivates our work in \cref{sec:motivation}.
Finally, we elaborate the proposed ClearCLIP for open-vocabulary semantic segmentation in \cref{sec:ClearCLIP}.

\subsection{Preliminary on CLIP}
\label{sec:preliminary}
\subsubsection{ViT architecture.}
\label{sec:vit}
A ViT-based CLIP model \cite{radford2021learning} consists of a series of residual attention blocks.
Each of these blocks takes as input a collection of visual tokens $X = [x_{\textup{cls}}, x_1, \dots, x_{h\times w}]^T$, where $x_{\textup{cls}}$ represents the global class token, and $\{x_i| i= 1, 2, \dots, h\times w\}$ denote local patch tokens.
For brevity, we omit the layer number and format a residual attention block as follows:
\begin{align}
    &q =\textup{Proj}_q(\textup{LN}(X)),\ k=\textup{Proj}_k(\textup{LN}(X)),\ v=\textup{Proj}_v(\textup{LN}(X)) \\
    &X_{\textup{sum}} = X_{\textup{res}} + X_{\textup{attn}} = X + \textup{Proj}(\textup{Attn}_{qk}\cdot v) \\
    &X = X_{\textup{sum}} + \textup{FFN}(\textup{LN}(X_{\textup{sum}})),
\end{align}
where LN denotes layer normalization, Proj represents a projection layer, and FFN stands for a feed-forward network. $X_{\textup{res}}$ and $X_{\textup{attn}}$ denote the residual connection and the attention output.
Additionally, $\textup{Attn}_{qk} = \textup{softmax}(\frac{qk^T}{\sqrt{d_k}})$ represents the $q$-$k$ attention, where $d_k$ is the dimension of $k$.

\subsubsection{Contrastive pre-training.}
CLIP employs a transformer-based visual encoder $\mathcal{V}$ and text encoder $\mathcal{T}$ to produce visual representations $X^{\textup{visual}}_{\textup{cls}}$ and text representations $X^{\textup{text}}$ for each image-text pair.
The pre-training of CLIP is grounded in the contrastive loss.
Given a batch of image-text pairs, CLIP is trained to maximize the cosine similarity between the visual representations $X^{\textup{visual}}_{\textup{cls}}$ and their corresponding text representations $X^{\textup{text}}$, while simultaneously minimizing the similarity of these representations from different pairs.

\subsubsection{Open-vocabulary dense inference.}
To adapt CLIP for open-vocabulary semantic segmentation, a baseline approach is to perform dense patch-level classification.
Given an image, the image encoder $\mathcal{V}$ is used to extract its visual representations $X^{\textup{visual}}=[x^{\textup{visual}}_{\textup{cls}}, X^{\textup{visual}}_{\textup{dense}}]^T$, where $X^{\textup{visual}}_{\textup{dense}}\in \mathbb{R}^{hw \times d}$ denotes the local patch representations in the $d$-dimensional latent space.
For the textual features, object labels with $C$ classes are firstly integrated into a prompt template ``\texttt{a photo of a \{label\}.}'' to obtain the text descriptions.
These descriptions are then fed into CLIP's text encoder to generate the text representations for all $C$ classes $X^{\textup{text}}\in \mathbb{R}^{C\times d}$.
The final segmentation map $\mathcal{M} \in \mathbb{R}^{hw\times 1}$ is computed as follows:
\begin{equation}
    \mathcal{M} = \mathop{\arg\max}_{c} \, \textup{cos}(X^{\textup{visual}}_{\textup{dense}}, X^{\textup{text}}). \label{eq:dense}
\end{equation}

\begin{figure}[t!]
  \centering
  \begin{subfigure}{0.5\linewidth}
    \includegraphics[width=0.9\linewidth]{ 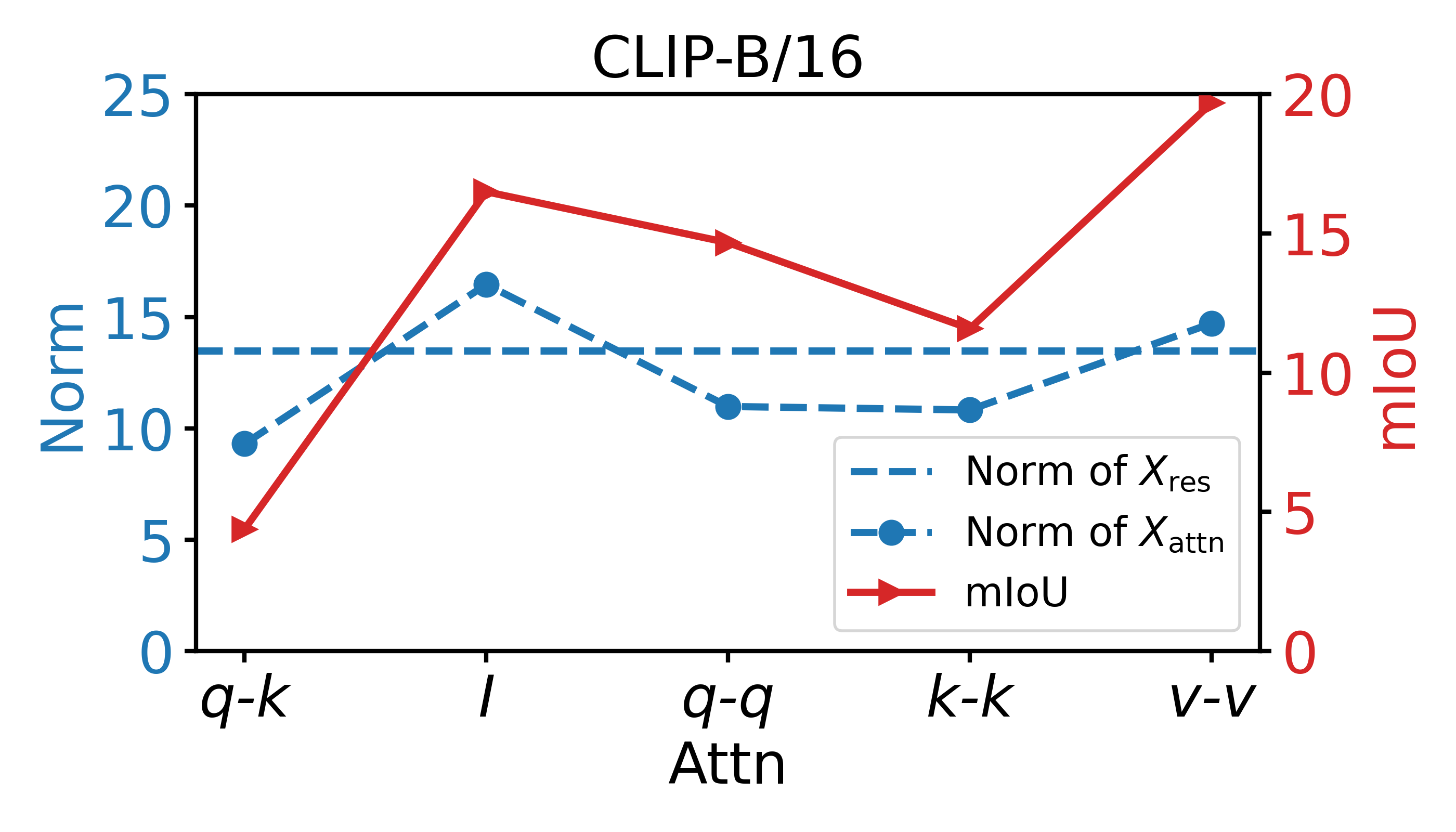}
  \end{subfigure}
  \begin{subfigure}{0.49\linewidth}
  \includegraphics[width=0.9\linewidth]{ 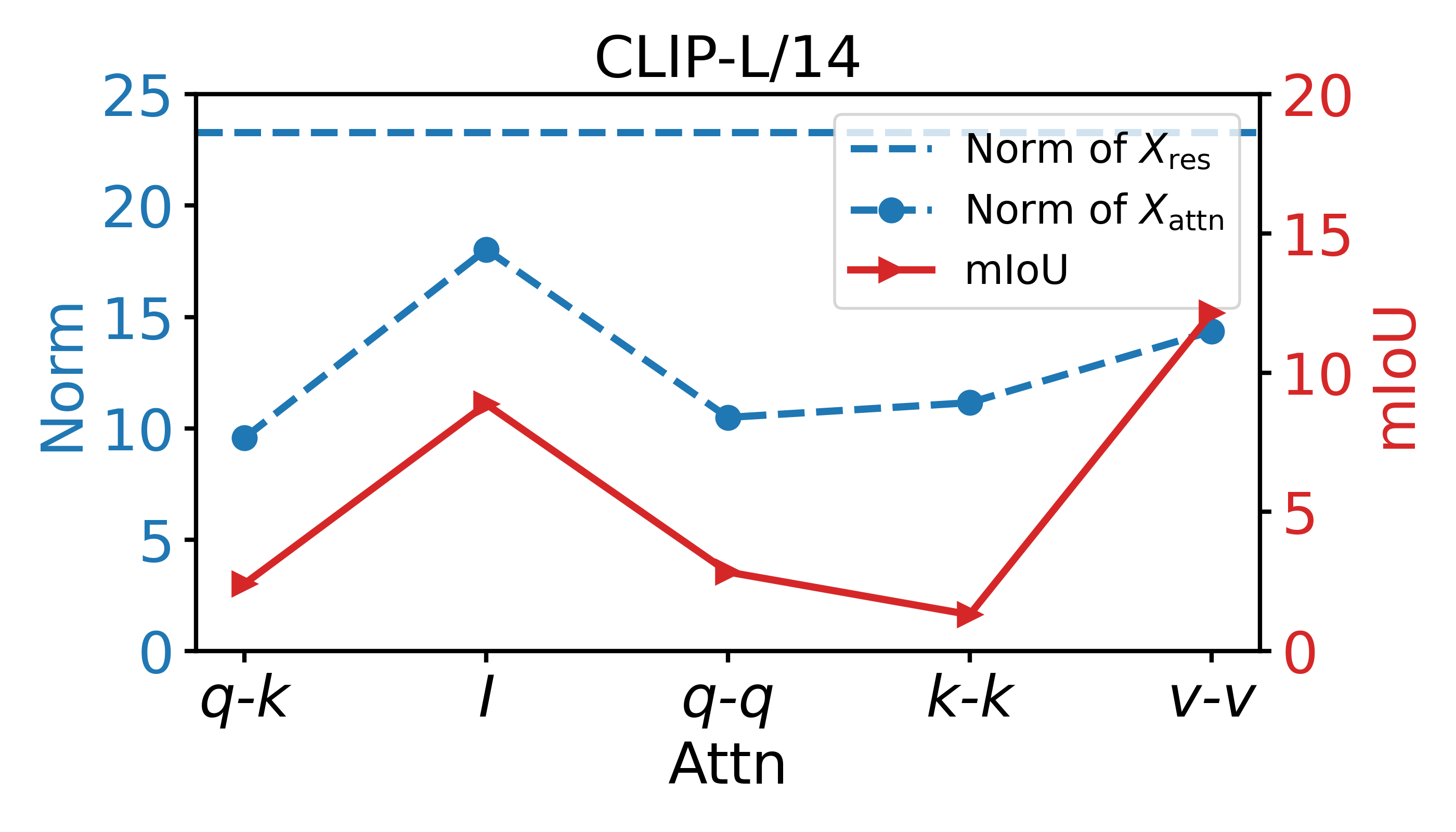}
  \end{subfigure}
  \caption{Comparison of norms and mIoU of different attention mechanisms for CLIP-B/16 (left) and CLIP-L/14 (right). The norm curve of $\xattn$ shows a positive correlation with the mIoU curve. A larger norm of $\xres$ in CLIP-L/14 impedes the enhancement of performance through the revision of attention mechanisms.}
  \label{fig:Norm_miou}
\end{figure}

\subsection{Motivation}
\label{sec:motivation}

The aforementioned baseline in \cref{eq:dense} often fails to achieve satisfactory results \cite{zhou2022extract}. This is probably because the CLIP is trained with image-level contrastive loss between vision and language, leading to poor alignment between local image regions and text representations \cite{wu2023clipself}.
Several studies \cite{zhou2022extract,li2023clip,bousselham2023grounding,wang2023sclip} have attempted to address this challenge with minimal modifications to CLIP without retraining.
At the core, they propose to revise the vanilla $\textup{Attn}_{qk}$ in the last self-attention layer to an identical attention \cite{zhou2022extract} or self-self attention \cite{li2023clip,bousselham2023grounding,wang2023sclip}, \ie, $\textup{Attn}_{qq}$, $\textup{Attn}_{kk}$ or $\textup{Attn}_{vv}$, aiming at re-organizing the spatial information.
As shown in \cref{fig:Norm_miou}, they successfully improve the baseline, with mIoU reaching up to nearly 20.0 from only 4.4 of CLIP with ViT-B/16 architecture (CLIP-B/16) on the COCOStuff dataset. 
However, there are still several important challenges. Firstly, previous works still generate sub-optimal results with noises in segmentation maps. Secondly, these methods fail to obtain reasonable results when using a larger model, such as ViT-L/14. In \cref{fig:Norm_miou}, $\textup{Attn}_{qq}$ and $\textup{Attn}_{kk}$ are even worse than the vanilla $\textup{Attn}_{qk}$ with more noises in segmentation maps. 
Such counter-intuitive phenomena indicates that existing works may have missed some important issues when adapting the CLIP model for dense prediction tasks. 
In this work, we are curious about \textit{where} and \textit{how} these noises in segmentation results originate and surface.

\subsection{ClearCLIP}
\label{sec:ClearCLIP}
As explained in \cref{sec:vit}, a block in the ViT-based CLIP contains three modules, \ie, the residual connection, the self-attention layer and the feed forward network. 
We delve into these modules to diagnose their effects on open-vocabulary semantic segmentation tasks. 
Finally, we propose ClearCLIP, a simple yet effective solution to produce clearer and more accurate segmentation maps.

\subsubsection{Residual connection.}
We begin our analysis by comparing the Frobenius norm of the residual connection $X_{\textup{res}}$ with different attention outputs $X_{\textup{attn}}$ at the last block from CLIP-B/16 and CLIP-L/14 models on the COCOStuff dataset.
As illustrated in \cref{fig:Norm_miou}, we can easily observe the commonalities and distinctions of these two sub-figures. The main commonality is that the mIoU curve and the norm curve of $\xattn$ exhibit a certain degree of positive correlation. 
The distinctions are: 1) the norm of $\xres$ in CLIP-B/16 is much smaller than that of CLIP-L/14; and 2) the attention modifications in CLIP-B/16 show consistent improvements over the \textit{q-k} baseline while those in CLIP-L/14 do not. 
Therefore, we hypothesize that the attention modification is effective only when the influence (or norm) of $\xres$ is minimal. 
In other words, $\xres$ substantially impairs the performance of the CLIP family on dense inference tasks.

\begin{figure}[t]
\begin{minipage}{0.7\textwidth}
\centering
\includegraphics[width=1.0\linewidth]{ 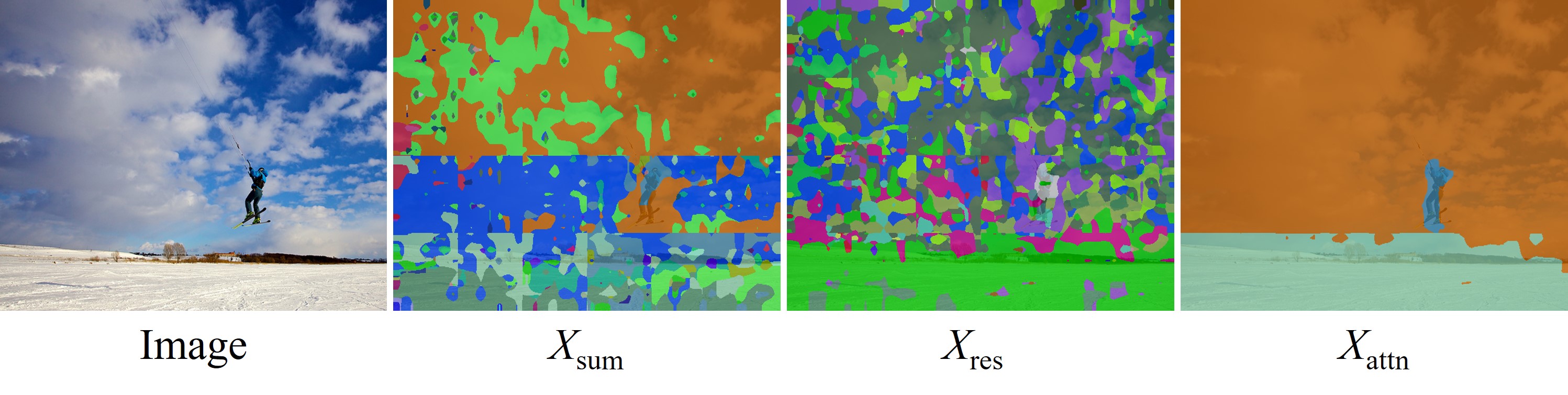}
\end{minipage}
\hfill
\begin{minipage}{0.29\textwidth}
	\centering
	\tabcolsep4pt
	\begin{tabular}{ll}
		\toprule
		Features & mIoU \\
		\midrule
        $X_{\textup{sum}}$ & 4.4  \\
        $X_{\textup{res}}$ & 0.01 \\
        $X_{\textup{attn}}$ & 11.6  \\
		\bottomrule
	\end{tabular}
\end{minipage}
\caption{Open-vocabulary semantic segmentation using different feature maps of CLIP-B/16 model on the COCOStuff dataset. A visualization of an example (left) and quantitative results (right).}
\label{fig_table}
\end{figure}

To investigate this hypothesis, we conduct open-vocabulary semantic segmentation experiments based on CLIP-B/16 using $X_{\textup{sum}}$, $X_{\textup{res}}$ and $X_{\textup{attn}}$.
Experimental results on the COCOStuff dataset are illustrated in \cref{fig_table}.
Surprisingly, we discover that the mIoU of $\xres$ is close to zero, suggesting that the residual connection may not be helpful for image segmentation. 
In contrast, $X_{\textup{attn}}$ alone could achieve much higher mIoU than $X_{\textup{sum}}$. 
The visualizations in \cref{fig_table} demonstrate that the noisy segmentation map of CLIP could be decomposed into a muddled map of $X_{\textup{res}}$ and a clearer map of $X_{\textup{attn}}$. 
According to these experimental results, we can primarily conclude that noises in segmentation maps mainly come from the residual connection.

\begin{figure}[tb]
  \centering
  \begin{subfigure}{0.32\linewidth}
    \includegraphics[width=1.0\linewidth]{ 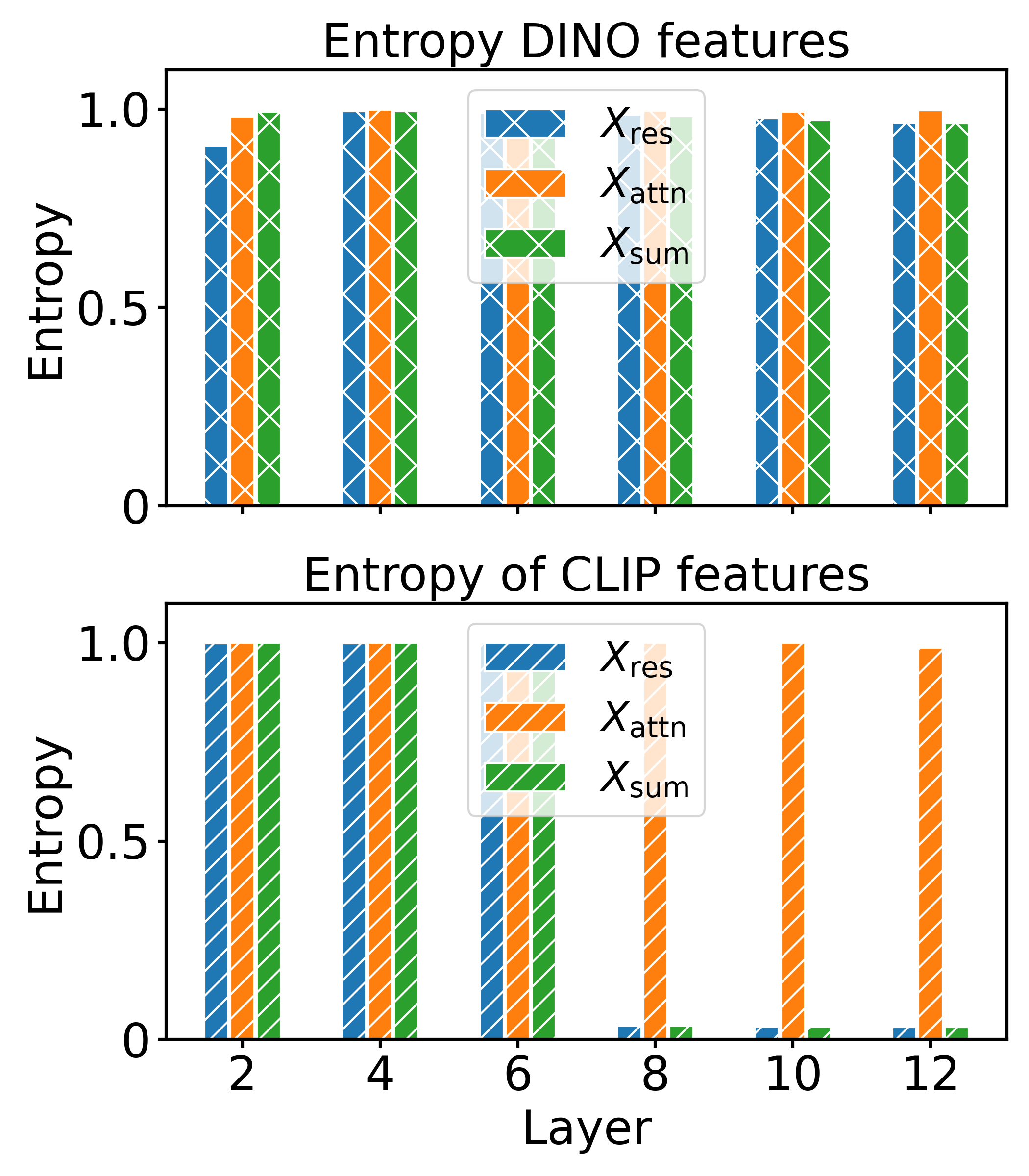}
    \caption{Entropy}
    \label{fig:entropy}
  \end{subfigure}
  \hfill
  \begin{subfigure}{0.32\linewidth}
  \includegraphics[width=1.0\linewidth]{ 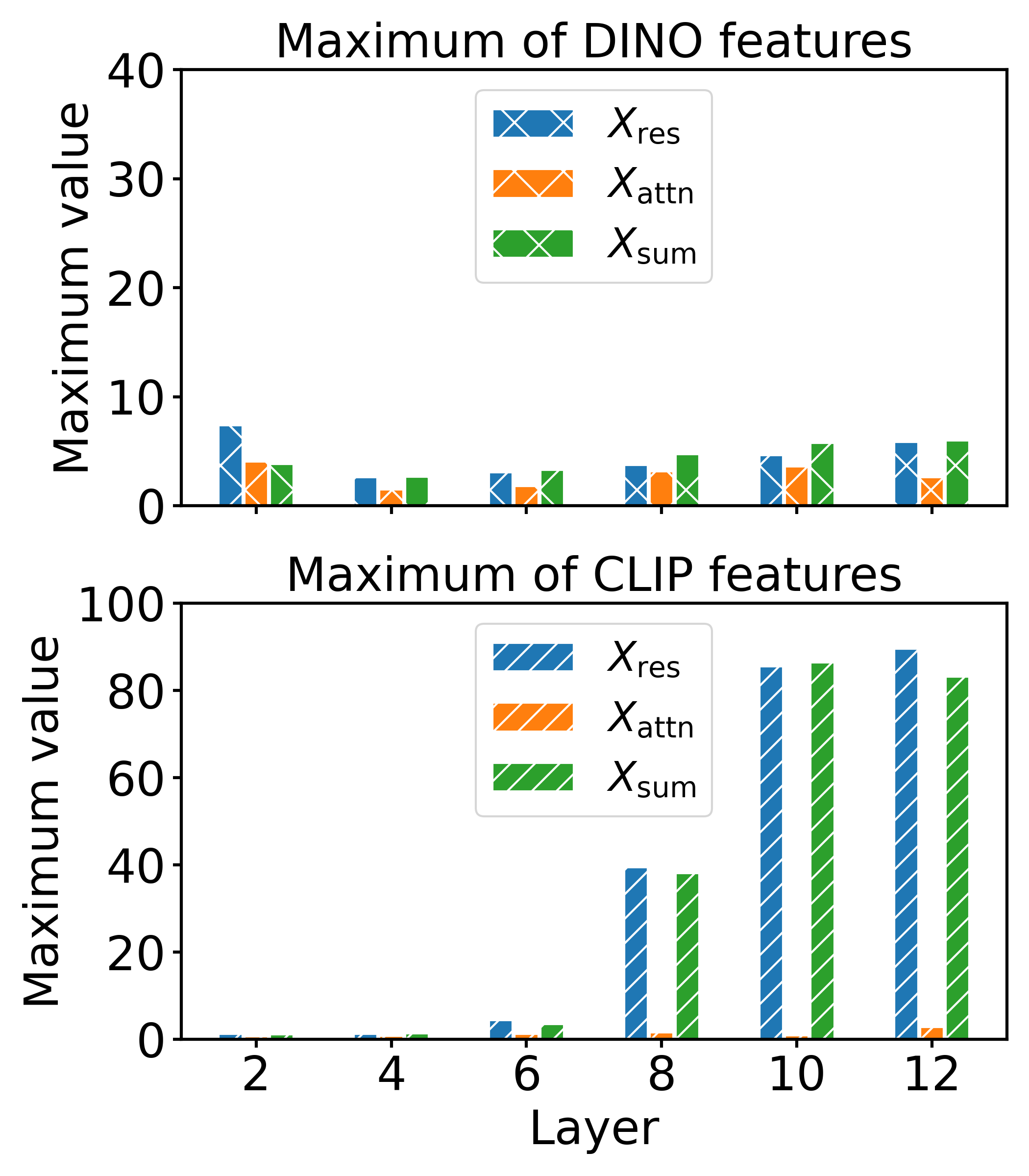}
    \caption{Maximum}
    \label{fig:maximum}
  \end{subfigure}
   \hfill
  \begin{subfigure}{0.32\linewidth}
   \includegraphics[width=1.0\linewidth]{ 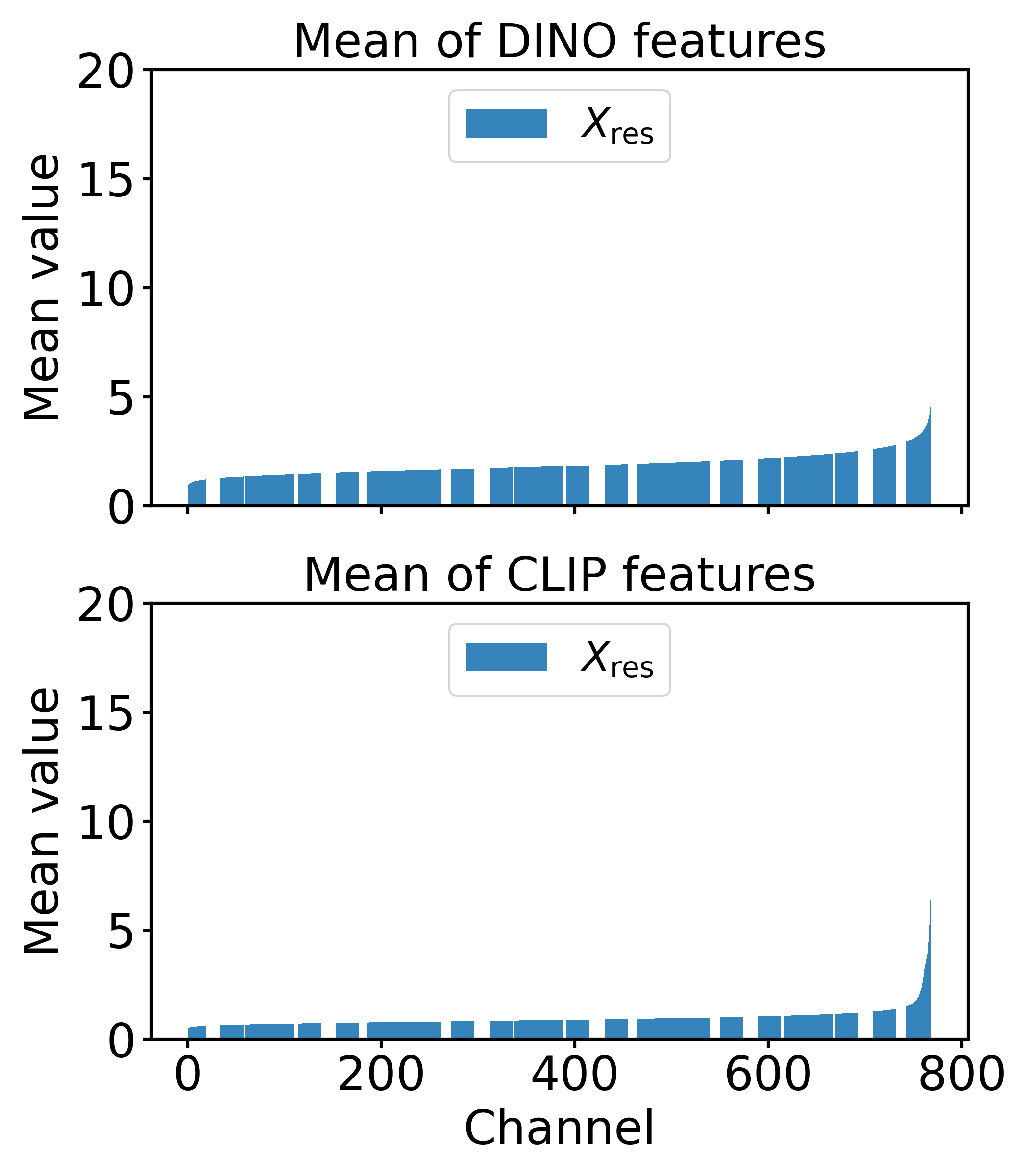}
    \caption{Mean}
    \label{fig:mean}
  \end{subfigure}
  \caption{Statistics of three feature maps for DINO-B/16 and CLIP-B/16.
  }
  \label{fig:statistics}
\end{figure}

To gain a deeper understanding of how these noises emerge in semantic segmentation tasks, we conduct a comparative analysis of feature statistics between CLIP-B/16 and DINO-B/16. 
The latter has demonstrated robust capabilities in learning transferable and semantically consistent dense features for various downstream tasks \cite{melas2022deep,hamilton2022unsupervised,lan2024smooseg}. 
We first compare the normalized entropies \cite{gray2011entropy} along layers, which is calculated by
\begin{align}
    H(X^L) = -\frac{1}{\textup{log}(hw\times d)}\sum_{i,j}p(X^L_{i,j})\,\textup{log}\,p(X^L_{i,j}) ,\ \ \ 
    p(X^L_{i,j}) = \frac{e^{X^L_{i,j}}}{\sum_{m,n}e^{X^L_{m,n}}}, \label{eq:entropy}
\end{align}
where $X^L$ denotes the feature map, \ie, $X_{\textup{sum}}$, $X_{\textup{res}}$ and $X_{\textup{attn}}$, at the $L$-th layer of the ViT network. 
As shown in \cref{fig:entropy}, we can see that the entropy of $X^L$ does not change much across layers for DINO-B/16. 
On the contrary, for CLIP-B/16, only the entropy of $\xattn$ remains the same across the layers, while the entropies of $\xsum$ and $\xres$ sharply decrease to near-zero. 
According to \cref{eq:entropy}, a low entropy indicates that there are a few peak values in $X^L$. 
Therefore, we examine the average maximum values of $\textup{max}_{i,j}X^L_{i,j}$ in \cref{fig:maximum}. 
For DINO-B/16, the maximum values of each type of feature maps remain relatively stable along layers, typically lower than 10, resulting in consistent entropies across different layers.
In contrast, for CLIP-B/16, the maximum values of $X_{\textup{res}}$ and $X_{\textup{sum}}$ gradually increase with the layer depth, peaking nearly 90 times higher at the last layer compared to earlier ones. 
Consequently, the entropies of $X_{\textup{res}}$ and $X_{\textup{sum}}$ sharply decline, approaching near-zero from the middle layers of ViT. 
Through visualizations of several feature maps (see supplementary material), we empirically found that these peak values appear in a few channels. 
To verify our observation, we calculate the average normalized mean values of each channel in $X_{\textup{res}}$ after sorting them in ascending order, and visualize them in \cref{fig:mean}. 
We can observe that a few channels dominate the peak values in $\xres$ which echoes our discovery from feature maps. 

Intuitively, these channel-wise statistics represent the global characteristics of $X^L$ since they are independent of local patterns. 
If $\xres$ and $\xsum$ are with low entropy and predominantly influenced by a few channels, it is highly probable that local information is being compromised. 
As depicted in \cref{eq:dense}, distinguishing between two feature vectors with cosine similarity becomes challenging if they share the same dominant channels.
While this characteristic is not harmful in itself for image recognition tasks that prioritize global information, it may result in sub-optimal performance when adapting CLIP to dense prediction tasks that emphasize local information.
Theoretically, this phenomenon becomes more pronounced in larger vision transformer models with deeper layers.
This analysis sheds light on why existing modifications on self-attention fail to yield satisfactory results when applied to the CLIP-L/14 model. 

To further demonstrate how $\xres$ affects the performance of CLIP, we introduce a scaling factor $\alpha$\footnote{SCLIP \cite{wang2023sclip} could be roughly regarded as a special case of $\alpha=2$, \ie, $\textup{Proj}((\textup{Attn}_{qq}+\textup{Attn}_{kk})\cdot v)\approx \textup{Proj}(2\textup{Attn}_{qk}\cdot v)\approx 2X_{\textup{attn}}$.}, $X_{\textup{sum}} = X_{\textup{res}} + \alpha X_{\textup{attn}}$, which controls the relative influence of $\xattn$ over $\xres$.
Our experimental results in \cref{fig:alpha} demonstrate that a larger $\alpha$ significantly enhances the performance, which clearly illustrates the adverse impact of $X_{\textup{res}}$ on the performance. 
Finally, we propose to directly discard the residual connection to achieve the best performance on dense vision-language inference tasks.

\subsubsection{Feed-forward network.}
The feed-forward network (FFN) in a transformer architecture plays a crucial role in modeling relationships and patterns within the data. 
However, recent work \cite{gandelsman2023interpreting} has revealed that the FFN has a negligible effect on image representation during the inference process. 
CLIPSurgery \cite{li2023clip} finds that the FFN features at the last attention block have a significantly larger cosine angle with the final classification feature, and therefore proposes to discard the FFN for dense prediction tasks.
In our work,
we empirically find that removing the FFN has minimal effect on open-vocabulary semantic segmentation tasks when applied to the vanilla CLIP model. 
However, as shown in \cref{fig:ablation}, when coupled with the removal of the residual connection, discarding the FFN leads to improved results, particularly with a larger model size. 
The rationale for this improvement is that removing the residual connection significantly alters the input to the FFN, consequently affecting its output. 
Therefore, removing the FFN output potentially mitigates its negative impact on performance.

\subsubsection{Our solution.}
Based on the above analysis, we propose a straightforward solution to adapt CLIP for open-vocabulary semantic segmentation. 
Specifically, we propose to use the attention output of the last self-attention layer\footnote{The final projection layer is omitted here for brevity.}
\begin{equation}
    X^{\textup{visual}} = X_{\textup{attn}} = \textup{Proj}(\textup{Attn}_{(\cdot) (\cdot)} \cdot v),
    \label{eq:solution}
\end{equation}
for vision-language inference. 
Inspired by previous works, we could use different combinations of \textit{query-key} in the attention mechanism $\textup{Attn}_{(\cdot) (\cdot)}$. 
In practice, we find that $\textup{Attn}_{qq}$ consistently achieves better performance in most cases and thus opt to use it by default.

\section{Experiments}
\label{sec:experiment}

\subsection{Experimental Setups}
\subsubsection{Datasets \& metric.}
Our solution is extensively evaluated on eight benchmark datasets widely employed for open-vocabulary semantic segmentation.
Following \cite{cha2023learning}, these datasets can be categorized into two groups: 1) with background category: 
PASCAL VOC \cite{everingham2012pascal} (\textbf{VOC21}),  PASCAL Context \cite{mottaghi2014role} (\textbf{Context60}) and COCO Object \cite{caesar2018coco} (\textbf{Object});
and 2) without background category:
PASCAL VOC20 \cite{everingham2012pascal} (\textbf{VOC20}), PASCAL Context59 \cite{mottaghi2014role} (\textbf{Context59}),  COCOStuff \cite{caesar2018coco} (\textbf{Stuff}), Cityscapes \cite{cordts2016cityscapes} and ADE20K \cite{zhou2019semantic}.

We utilize the implementations provided by MMSegmentation \cite{contributors2020mmsegmentation}, employ a sliding window strategy, and resize input images to have a shorter side of 448 pixels.
Following established practices, we avoid text expansions of class names and rely solely on the standard ImageNet prompts \cite{radford2021learning}. 
For a fair comparison, no post-processing is applied to any of the methods evaluated. 
\textit{Our method does not need any retraining or fine-tuning}. 
Therefore, we can directly evaluate its performance on the validation set of all datasets.
For evaluating semantic segmentation tasks, we employ the mean Intersection over Union (mIoU) metric.

\subsubsection{Baselines.}
We compare our method with two types of open-vocabulary semantic segmentation methods:
1) \textbf{training-free} methods including CLIP \cite{radford2021learning}, MaskCLIP \cite{zhou2022extract}, ReCo \cite{shin2022reco}, CLIPSurgery \cite{li2023clip}, GEM \cite{bousselham2023grounding}, and SCLIP \cite{wang2023sclip};
and 2) \textbf{weakly-supervised} methods including GroupViT \cite{xu2022groupvit}, SegCLIP \cite{luo2023segclip}, OVSegmentor \cite{xu2023learning}, 
PGSeg \cite{zhang2023uncovering}, ViewCo \cite{ren2023viewco}, CoCu \cite{xing2023rewrite}, and TCL \cite{cha2023learning}.
Unless explicitly mentioned, all reported results are directly cited from the respective papers.
Additionally, we include results of the baselines based on CLIP-L/14 using our implementation for comprehensive evaluation.

\begin{table}[tb]
  \caption{Ablation results based on CLIP-B/16 architecture on five datasets \textit{without} background class. RC denotes the residual connection.
  }
  \tabcolsep5pt
  \label{tab:results_ablation}
  \centering
  \begin{tabular}{ccc|ccccc|c}
    \toprule
    Attn & RC & FFN & VOC20 & Context59 & Stuff & Cityscapes & ADE20k & Avg. \\
    \midrule
    $q$-$q$ & \ding{51} & \ding{51} & 68.4 & 24.9 & 14.7 & 20.8 & 7.6 & 27.3 \\
    $q$-$q$ & \ding{51} & \ding{55} & 62.8 & 25.5 & 14.6 & 19.5 & 6.9 & 25.9 \\
    $q$-$q$ & \ding{55} & \ding{51} & 77.6 & 31.8 & 21.0 & 23.4 & 14.7 & 33.7 \\
    $q$-$q$ & \ding{55} & \ding{55} & \textbf{80.9} & \textbf{35.9} & \textbf{23.9} & \textbf{30.0} & \textbf{16.7} & \textbf{37.5} \\
  \bottomrule
  \end{tabular}
\end{table}

\subsection{Analysis and Discussion}
In this section, we present comprehensive experiments to validate the effectiveness of our solution.
To ensure a rigorous comparison, 
our experiments primarily focus on five datasets without the background class. 

\subsubsection{Ablation study.}
We conduct ablation studies using the CLIP-B/16 model to assess the effectiveness of our solution.
The results are summarized in \cref{tab:results_ablation}.
Notably, the removal of the residual connection yields a significant performance improvement, increasing the average mIoU from 27.3 to 33.7. 
This result corroborates our assertion that residual features contain less local information, thereby influencing dense patch prediction.
Interestingly, removing the FFN alone does not yield better results. 
However, the model without both residual connection and FFN together achieves the best performance, with an mIoU of 37.5. 
This observation is reasonable since removing the residual connection alters the input to the FFN, consequently affecting its output. In this case, removing FFN potentially mitigates the negative impact on performance.

\begin{figure}[tb]
  \centering
  \begin{minipage}{0.64\linewidth}
    \centering
    \begin{subfigure}{0.49\linewidth}
      \includegraphics[width=1.0\linewidth]{ 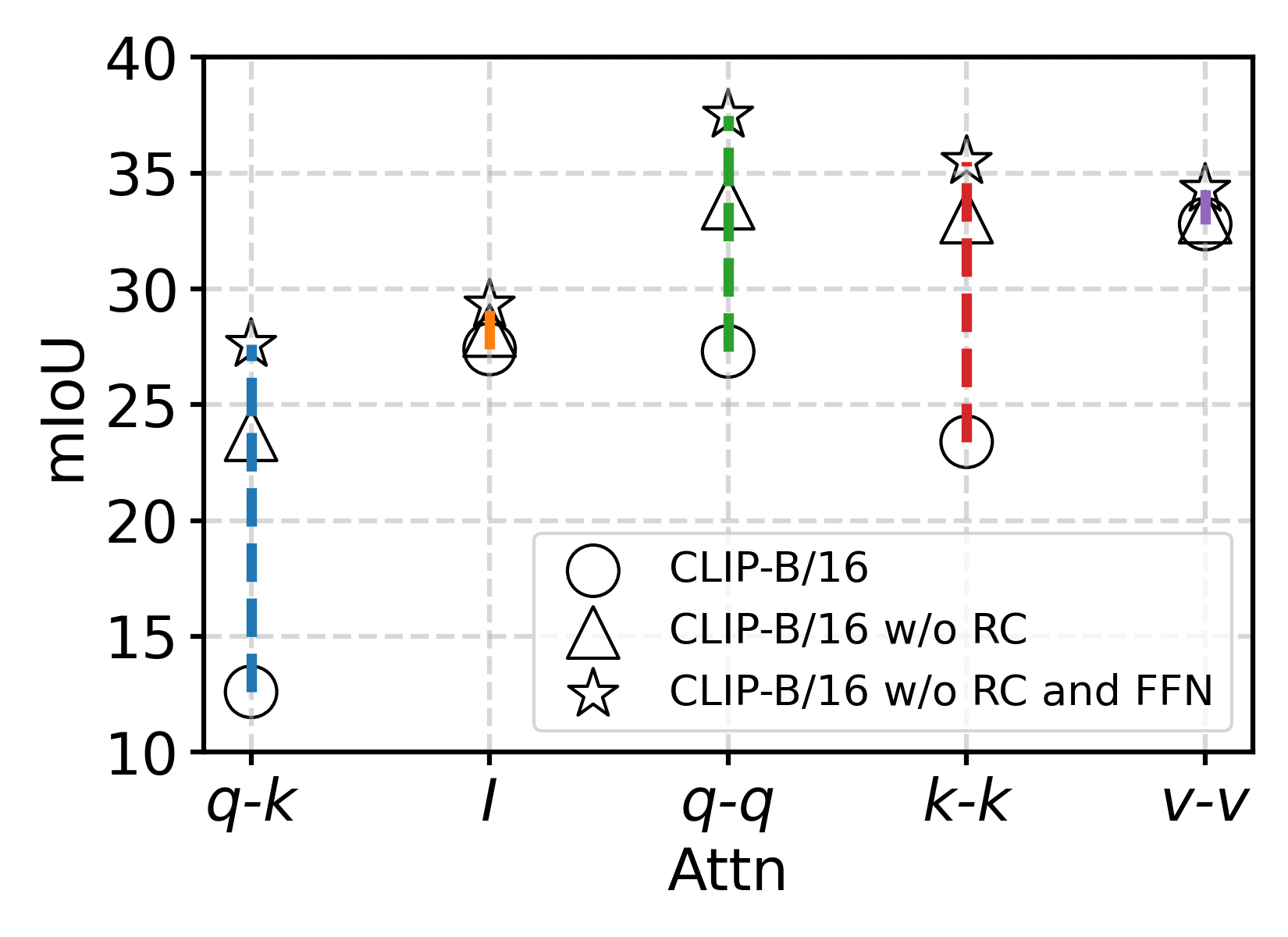}
      \caption{CLIP-B/16}
      \label{fig:ablation_clip_b16}
    \end{subfigure}
    \begin{subfigure}{0.49\linewidth}
      \includegraphics[width=1.0\linewidth]{ 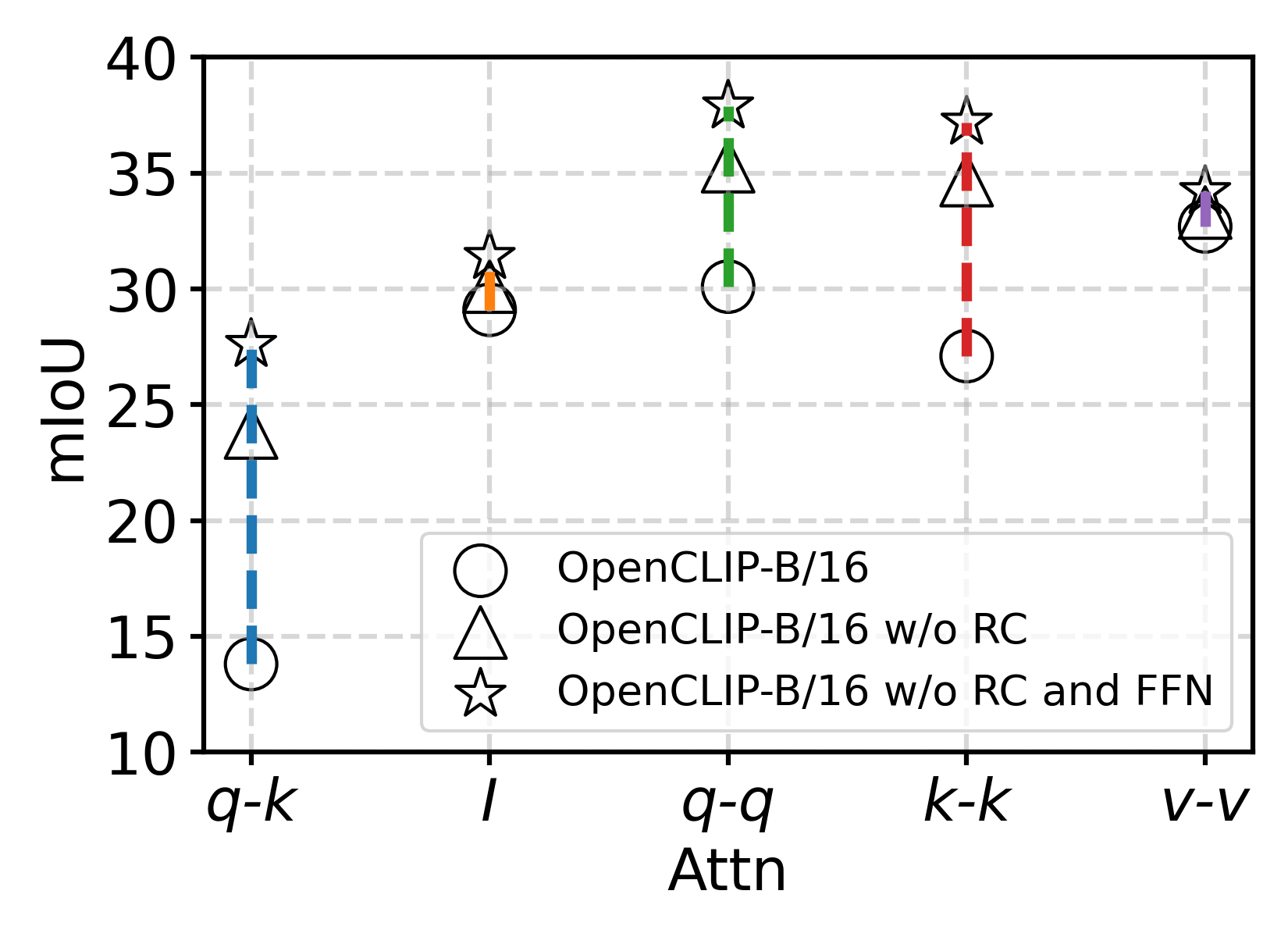}
      \caption{OpenCLIP-B/16}
      \label{fig:ablation_openclip_b16}
    \end{subfigure}
    \begin{subfigure}{0.49\linewidth}
      \includegraphics[width=1.0\linewidth]{ 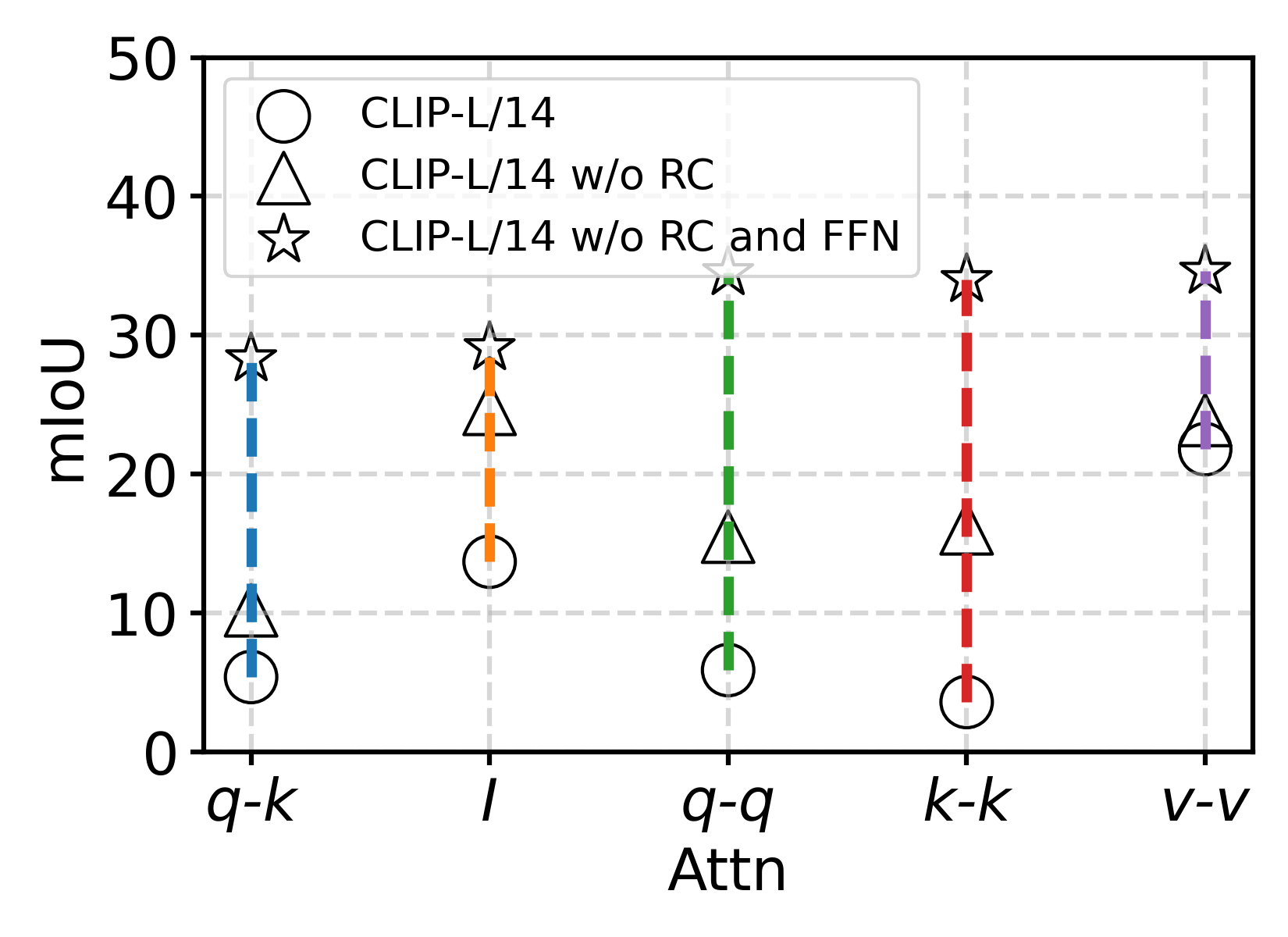}
      \caption{CLIP-L/14}
      \label{fig:ablation_clip_L14}
    \end{subfigure}
    \begin{subfigure}{0.49\linewidth}
      \includegraphics[width=1.0\linewidth]{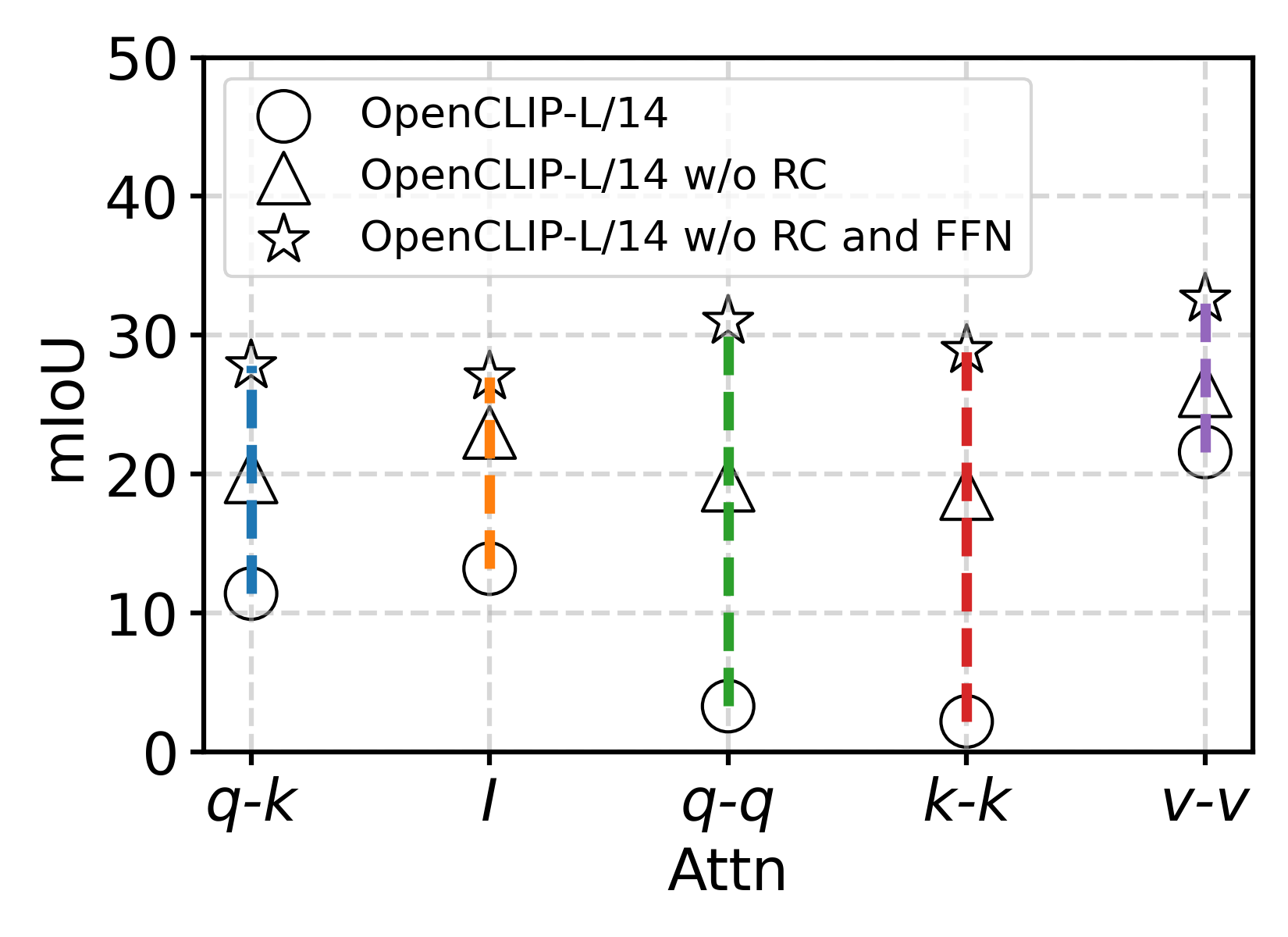}
      \caption{OpenCLIP-L/14}
      \label{fig:ablation_opencip_L14}
    \end{subfigure}
    \caption{Ablation study on different architectures and different attention mechanisms.}
    \label{fig:ablation}
  \end{minipage}
  \begin{minipage}{0.35\linewidth}
    \centering
    \begin{subfigure}{0.9\linewidth}
      \includegraphics[width=1.0\linewidth]{ 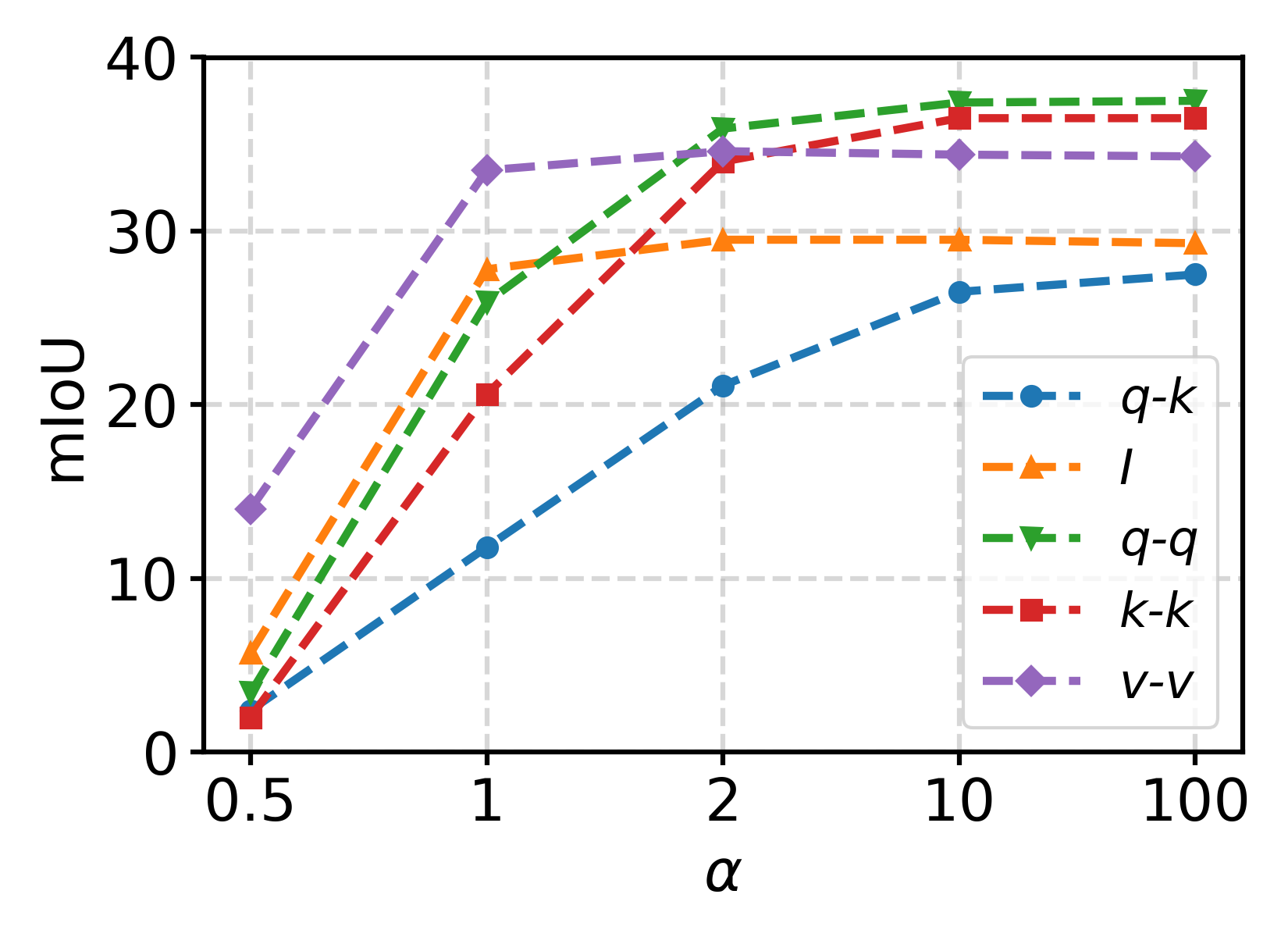}
      \caption{CLIP-B/16}
      \label{fig:alpha_b16}
    \end{subfigure}
    \begin{subfigure}{0.9\linewidth}
      \includegraphics[width=1.0\linewidth]{ 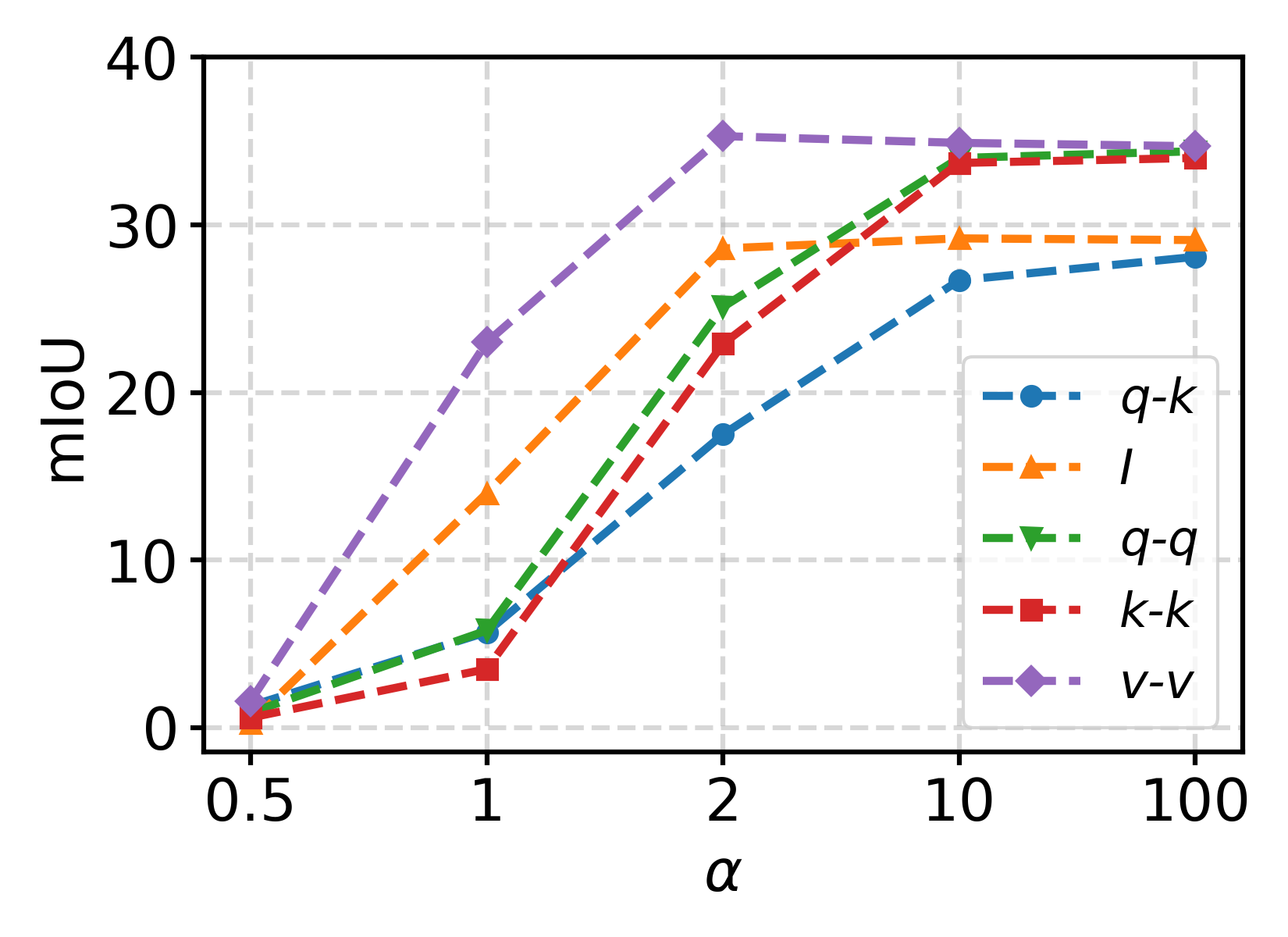}
      \caption{CLIP-L/14}
      \label{fig:alpha_l14}
    \end{subfigure}
    \caption{Segmentation results w.r.t. the scaling factor $\alpha$.}
    \label{fig:alpha}
  \end{minipage}
\end{figure}

\subsubsection{Different architectures.}
Given the simplicity of our proposed solution, it could be seamlessly applied to different architectures. 
We conduct experiments with CLIP \cite{radford2021learning} and OpenCLIP \cite{cherti2023reproducible} using ViT-B/16 and ViT-L/14 models.
The results regarding the average mIoU on five datasets are depicted in \cref{fig:ablation}.
Our analysis reveals several noteworthy findings:
1) Across different architectures, the consistent achievement of superior segmentation results aligns with the removal of both the residual connection and FFN at the last transformer block. This emphasizes the effectiveness of our solution to adapt vision-language pre-training models for downstream tasks.
2) Notably, the self-self attention consistently outperforms the vanilla $q$-$k$ attention on our solution. For instance, in the CLIP-B/16 w/o RC and FFN model, the $q$-$q$ attention yields an average mIoU of 37.5, surpassing the 27.6 mIoU achieved by the $q$-$k$ attention.
3) For CLIP-L/14 and OpenCLIP-L/14 models, we observe that the self-self attention fails, and the performance of $q$-$q$ and $k$-$k$ attentions even falls below that of the vanilla $q$-$k$ attention. 
This highlights that existing works aiming at revising the attention mechanism do not address the core problem when adapting CLIP for open vocabulary semantic segmentation. 
In contrast, our solution of using the attention output leads to significant performance improvements.
4) Interestingly, for models with ViT-B/16 architecture, the improvement of our solution is less pronounced with the identical attention ($\id$) and the $v$-$v$ attention compared to other attention types. 
This phenomenon can be attributed to the fact that the $\id$ and $v$-$v$ attentions tend to sharpen the attention output, thereby increasing the norm of the attention output, as illustrated in \cref{fig:Norm_miou}. 
We assert that the enhancement of CLIP-B/16 and OpenCLIP-B/16 under the $\id$ and $v$-$v$ attentions primarily stems from implicitly eliminating the negative effect of the residual connection. 
Consequently, explicitly removing the residual connection and FFN leads to limited improvement. 
However, for models with ViT-L/14 architecture, where the norm of the residual connection is substantially larger, removing both the residual connection and FFN results in significant improvement.

\begin{table}[!t]
  \caption{Open-vocabulary semantic segmentation quantitative comparison on datasets \textit{without} a background class. $^\dagger$ denotes results directly cited from TCL \cite{cha2023learning}. SCLIP$^*$ denotes our reproduced results under the standard setting without class re-name tricks.
  }
  \tabcolsep3pt
  \label{tab:results_wo_background}
  \centering
  \begin{tabular}{lccccccc}
    \toprule
    Methods & Encoder & VOC20 & Context59 & Stuff & Cityscape & ADE20k & Avg. \\
    \midrule
    GroupViT$^\dagger$ \cite{xu2022groupvit} & ViT-S/16 & 79.7 & 23.4 & 15.3 & 11.1 & 9.2 & 27.7\\
    CoCu \cite{xing2023rewrite} & ViT-S/16 & - & - & 13.6 & 15.0 & 11.1 & - \\
    TCL \cite{cha2023learning} & ViT-B/16 & 77.5 & 30.3 & 19.6 & 23.1 & 14.9 & 33.1\\
    \midrule
    CLIP \cite{radford2021learning} & ViT-B/16 & 41.8 & 9.2 & 4.4 & 5.5 & 2.1 & 12.6\\
    MaskCLIP$^\dagger$ \cite{zhou2022extract} & ViT-B/16 & 74.9 & 26.4 & 16.4 & 12.6 & 9.8 & 28.0\\
    ReCo$^\dagger$ \cite{shin2022reco} & ViT-B/16 & 57.7 & 22.3 & 14.8 & 21.1 & 11.2 & 25.4\\
    CLIPSurgery \cite{li2023clip} & ViT-B/16 & - & - & 21.9 & \textbf{31.4} & - & - \\
    \textcolor{lightgray}{SCLIP} \cite{wang2023sclip} & \textcolor{lightgray}{ViT-B/16} & \textcolor{lightgray}{80.4} & \textcolor{lightgray}{34.2} & \textcolor{lightgray}{22.4} & \textcolor{lightgray}{32.2} & \textcolor{lightgray}{16.1} & \textcolor{lightgray}{37.1}\\
    SCLIP$^*$ \cite{wang2023sclip} & ViT-B/16 & 78.2 & 33.0 & 21.1 & 29.1 & 14.6 & 35.2\\
    \rowcolor{darkgray} ClearCLIP & ViT-B/16 & \textbf{80.9} & \textbf{35.9} & \textbf{23.9} & 30.0 & \textbf{16.7} & \textbf{37.5}\\
    \midrule
    CLIP \cite{radford2021learning} & ViT-L/14 & 15.8 & 4.5 & 2.4 & 2.9 & 1.2 & 5.4 \\
    MaskCLIP \cite{zhou2022extract} & ViT-L/14 & 30.1 & 12.6 & 8.9 & 10.1 & 6.9 & 13.7 \\
    SCLIP \cite{wang2023sclip} & ViT-L/14 & 60.3 & 20.5 & 13.1 & 17.0 & 7.1 & 23.6 \\
    \rowcolor{darkgray} ClearCLIP & ViT-L/14 & 80.0 & 29.6 & 19.9 & 27.9 & 15.0 & 34.5\\
  \bottomrule
  \end{tabular}
\end{table}

\subsubsection{Effect of amplifying the norm of attention output.}
To further explore the relationship between the residual connection and the attention output in open vocabulary semantic segmentation tasks, we conduct experiments using $\alpha=\{0.1, 1, 2, 10, 100\}$, explicitly amplifying the F-norm of $X_{\textup{attn}}$ to $\alpha^2$ times.
As shown in \cref{fig:alpha},
our results reveal a clear trend: as the scaling factor $\alpha$ increases, models with all types of attention exhibit significantly improved performance. 
As expected, performance sharply declines when $\alpha$ decreases from 1 to 0.5. 
These findings underscore the importance of enlarging the norm of the attention output to mitigate the negative effects of the residual connection, ultimately leading to substantially improved performance. 
Hence, our solution of removing the residual connection proves to be simple yet effective.
Additionally, these insights help elucidate the superior performance of SCLIP \cite{wang2023sclip}, which adopts the $q$-$q$ plus $k$-$k$ attention, as this attention mechanism roughly doubles the vanilla attention.

\begin{table}[t]
  \caption{Open-vocabulary semantic segmentation quantitative comparison on datasets \textit{with} a background class. $^\dagger$ denotes results directly cited from TCL \cite{cha2023learning}. SCLIP$^*$ denotes our reproduced results under the standard setting without class re-name tricks.
  }
  \tabcolsep8pt
  \label{tab:results_w_background}
  \centering
  \begin{tabular}{lccccc}
    \toprule
    Methods & Encoder & VOC21 & Context60 & Object & Avg. \\
    \midrule
    GroupViT$^\dagger$ \cite{xu2022groupvit} & ViT-S/16 & 50.4 & 18.7 & 27.5 & 32.2 \\
    SegCLIP \cite{luo2023segclip} & ViT-S/16 & 52.6 & 24.7 & 26.5 & 34.6 \\
    OVSegmentor \cite{xu2023learning} & ViT-B/16 & \textbf{53.8} & 20.4 & 25.1 & 33.1 \\
    PGSeg \cite{zhang2023uncovering} & ViT-S/16 & 53.2 & 23.8 & 28.7 & 35.2 \\
    ViewCo \cite{ren2023viewco} & ViT-S/16 & 52.4 & 23.0 & 23.5 & 33.0\\
    CoCu \cite{xing2023rewrite} & ViT-S/16 & 40.9 & 21.2 & 20.3 & 27.5\\
    TCL \cite{cha2023learning} & ViT-B/16 & 51.2 & 24.3 & 30.4 & 35.3 \\
    \midrule
    CLIP \cite{radford2021learning} & ViT-B/16 & 16.2 & 7.7 & 5.5 & 9.8 \\
    MaskCLIP$^\dagger$ \cite{zhou2022extract} & ViT-B/16 &  38.8 & 23.6 & 20.6 & 27.7 \\
    ReCo$^\dagger$ \cite{shin2022reco} & ViT-B/16 & 25.1 & 19.9 & 15.7 & 20.2 \\
    CLIPSurgery \cite{li2023clip} & ViT-B/16 & - & 29.3 & - & - \\
    GEM \cite{bousselham2023grounding} & ViT-B/16 & 46.2 & \textbf{32.6} & - & - \\
    \textcolor{lightgray}{SCLIP} \cite{wang2023sclip} & \textcolor{lightgray}{ViT-B/16} & \textcolor{lightgray}{59.1} & \textcolor{lightgray}{30.4} & \textcolor{lightgray}{30.5} & \textcolor{lightgray}{40.0} \\
    SCLIP$^*$ \cite{wang2023sclip} & ViT-B/16 & 51.4 & 30.5 & 30.0 & 37.3 \\
    \rowcolor{darkgray} ClearCLIP & ViT-B/16 & 51.8 & \textbf{32.6} & \textbf{33.0} & \textbf{39.1}\\
  \bottomrule
  \end{tabular}
\end{table}

\subsection{Comparison to State-of-the-art}

\subsubsection{Quantitative results.}
\cref{tab:results_wo_background} summarizes the performance of various open-vocabulary semantic segmentation models on datasets without a background class. We observe that our method ClearCLIP achieves the best results on four out of five datasets. 
ClearCLIP significantly outperforms TCL on all datasets, with an average improvement of 4.4 mIoU. 
We note that SCLIP also achieves much better performance compared to other methods. 
This is because SCLIP implicitly attenuates the residual connection by using the $q$-$q$ plus $k$-$k$ attention, roughly doubling the attention output. 
However, our ClearCLIP explicitly removes the residual connection and FFN, resulting in an average improvement of 3.3 mIoU over SCLIP. 
Interestingly, when adopting the ViT-L/14 model, both MaskCLIP and SCLIP fail to achieve satisfactory results, while our method obtains higher results, at 34.5 mIoU, much better than the 23.6 mIoU of SCLIP. 
Although this result is not better than those achieved with the ViT-B/16 model, it still demonstrates the better generality of ClearCLIP with different backbones.

We report the results on three datasets with a background class on \cref{tab:results_w_background}.
The performance of ClearCLIP is significantly better than all weakly-supervised state-of-the-art methods, with an average improvement of 3.8 mIoU over TCL. Additionally, ClearCLIP outperforms SCLIP, with performance improvements of 0.4, 2.1, and 3.0 mIoU on VOC21, Context60, and COCO Object datasets, respectively. These results fully demonstrate the effectiveness of our solution of decomposing CLIP's features for open-vocabulary semantic segmentation.

\begin{figure}[t]
    \centering
    \includegraphics[width=0.9\linewidth]{ 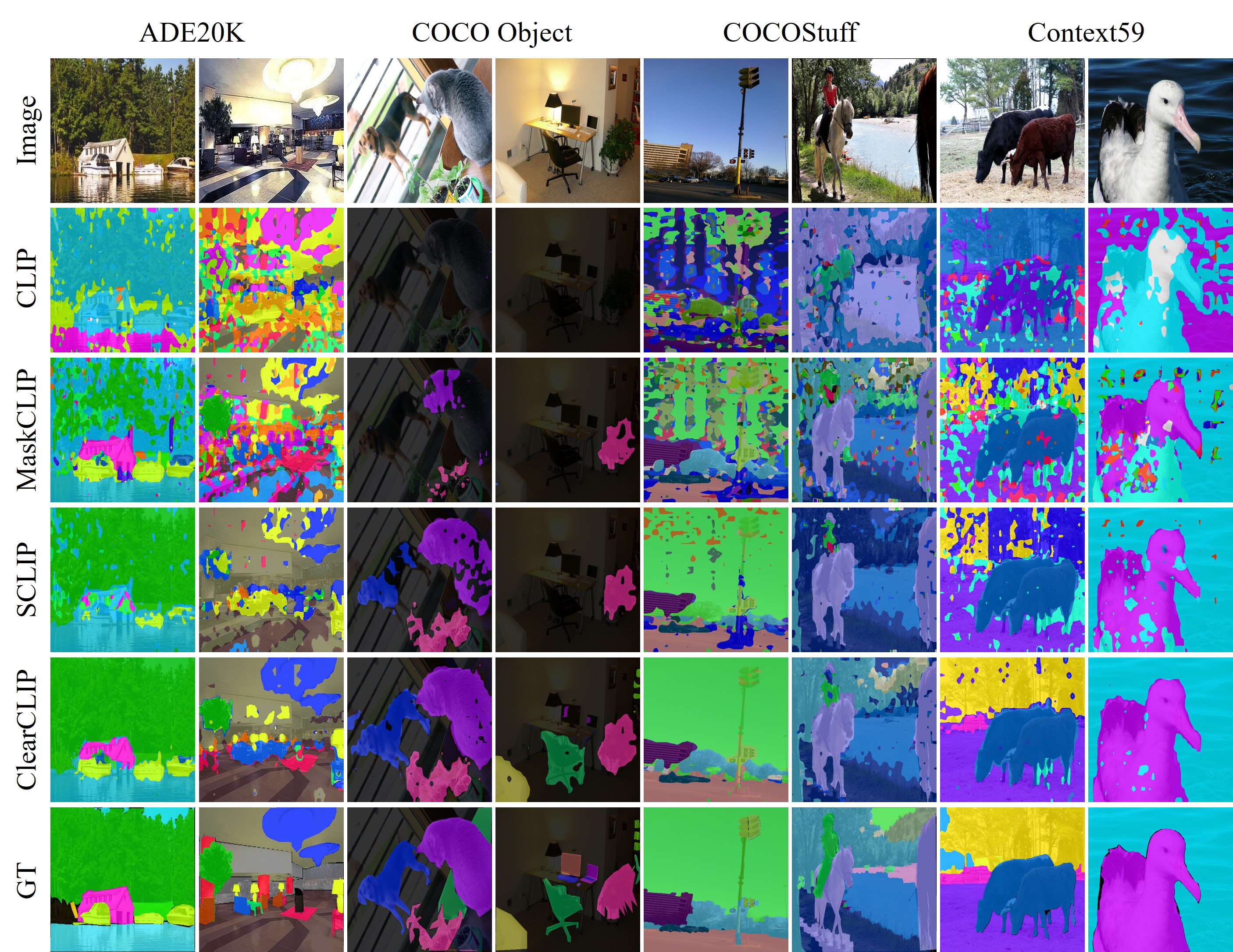}
    \caption{Qualitative comparison between open-vocabulary segmentation methods.}
    \label{fig:visualization}
\end{figure}

\subsubsection{Qualitative results.}
In \cref{fig:visualization}, we present a qualitative comparison between ClearCLIP and three training-free methods, \ie, CLIP, MaskCLIP, and SCLIP. 
Our observations are summarized as follows:
1) MaskCLIP exhibits good localization ability compared to CLIP but still generates segmentation maps with noticeable noise and many incoherent segments (e.g., those depicting a dog, cat, and duck in the 3rd and 8th columns);
2) SCLIP showcases the capability of detecting detailed semantic features with less noise compared to MaskCLIP;
3) ClearCLIP consistently produces much clearer and more accurate fine-grained segmentation maps than the other methods evaluated.
These observations validate our attempt to enhance the performance of open-vocabulary semantic segmentation by generating clearer segmentation maps through CLIP's representation decomposition.

\section{Conclusion}

In this study, we explore the origins and mechanisms of noisy segmentation results when utilizing the CLIP family for open-vocabulary semantic segmentation. 
We re-examine the architecture of CLIP and conduct a comparative analysis of feature statistics within the residual connection and the attention output. 
By investigating the differences in norm values across varied sizes of CLIP backbones, we discover that the residual connection serves as the primary source of segmentation noise. Additionally, through a comparative study between CLIP and DINO, we propose that the lack of local information in residual features stems from high-level supervision, which prioritizes global direction. 
Finally, we introduce ClearCLIP, a simple yet effective solution that removes the residual connection, adopts the self-self attention, and discards the FFN. ClearCLIP demonstrates superior performance and generalizability within the CLIP family.

\noindent \textbf{Acknowledgments.} 
This study is supported under the RIE2020 Industry Alignment Fund – Industry Collaboration Projects (IAF-ICP) Funding Initiative, as well as cash and in-kind contribution from the industry partner(s).

%
%
\bibliographystyle{splncs04}
\bibliography{main}

\newpage
\appendix
\section*{Appendix}
\addcontentsline{toc}{section}{Appendix}

\section{Ablation study with different backbones and datasets}
\label{sec:Different architectures}
\begin{figure}[h]
  \centering
  \begin{subfigure}{0.99\linewidth}
    \includegraphics[width=1.0\linewidth]{ 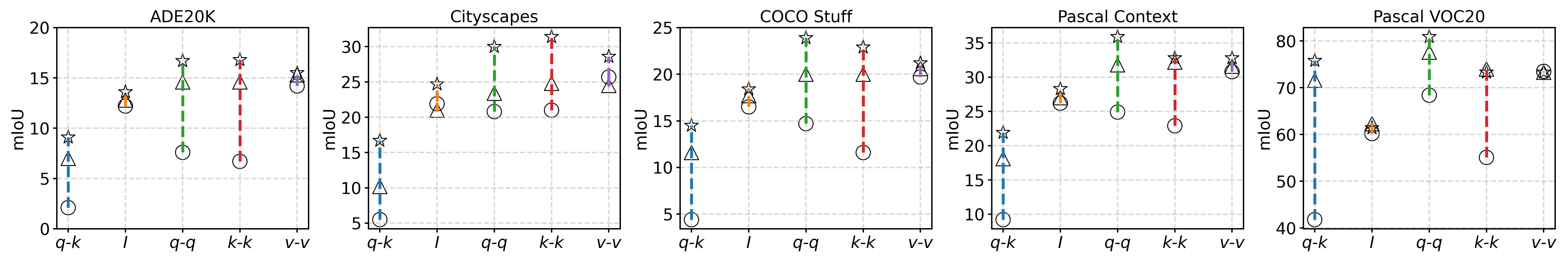}
    \caption{CLIP-B/16}
    \label{fig:ablation_clip_b16_supp}
  \end{subfigure}
  \begin{subfigure}{0.99\linewidth}
  \includegraphics[width=1.0\linewidth]{ 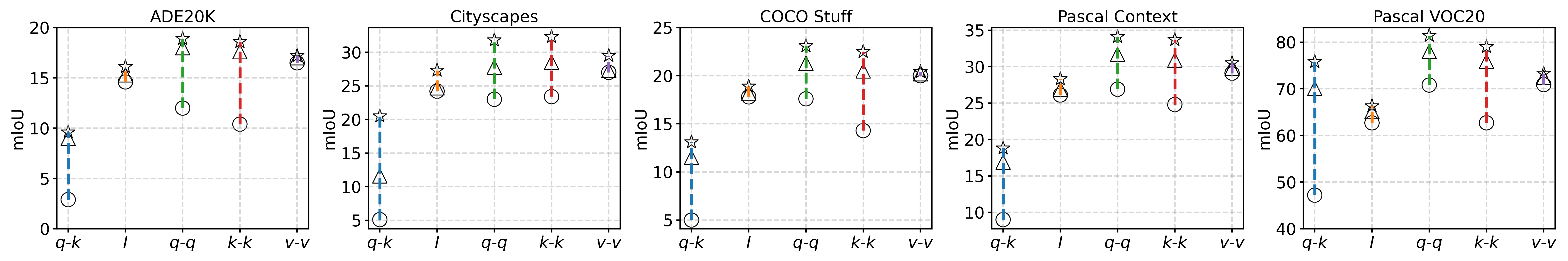}
    \caption{OpenCLIP-B/16}
    \label{fig:ablation_openclip_b16_supp}
  \end{subfigure}
  \begin{subfigure}{0.99\linewidth}
    \includegraphics[width=1.0\linewidth]{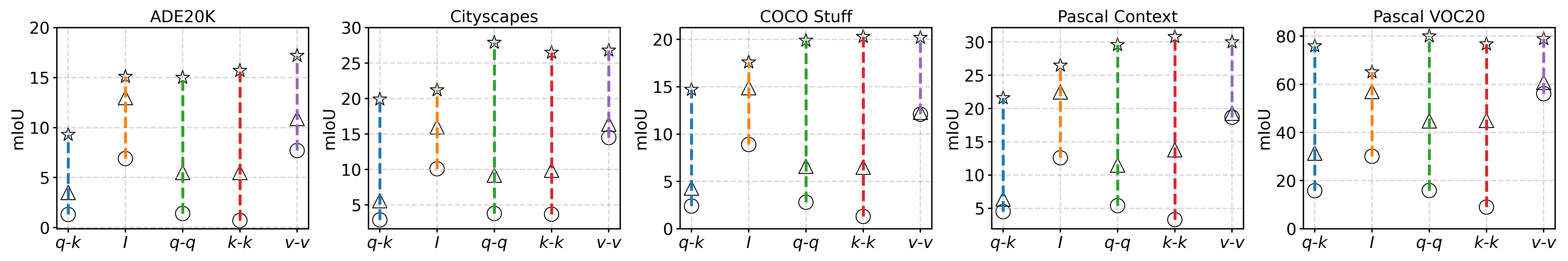}
    \caption{CLIP-L/14}
    \label{fig:ablation_clip_L14_supp}
  \end{subfigure}
  \begin{subfigure}{0.99\linewidth}
  \includegraphics[width=1.0\linewidth]{ 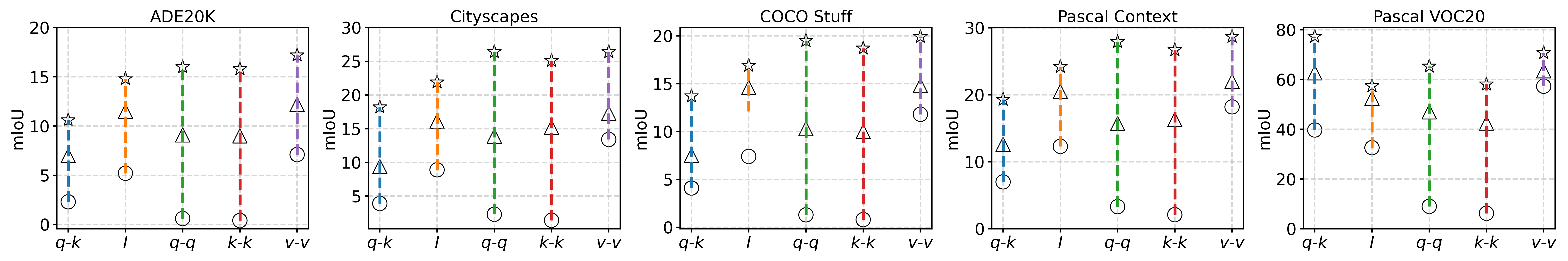}
    \caption{OpenCLIP-L/14}
    \label{fig:ablation_opencip_L14_supp}
  \end{subfigure}
  \caption{Ablation study on each dataset under different architectures and attention mechanisms. $\bigcirc$: original CLIP; $\bigtriangleup$: CLIP w/o residual connection; \ding{73}: CLIP w/o residual connection and FFN.}
  \label{fig:ablation_supp}
\end{figure}

We showcase the results of the ablation study for each dataset across different CLIP models in \cref{fig:ablation_supp}. It's clear that our method, which involves removing the residual connection and FFN, markedly enhances the open-vocabulary semantic segmentation capability of CLIP throughout all datasets. This enhancement is especially pronounced within the ViT-L/14 architecture, characterized by a larger norm of residual connection. These findings conclusively affirm the efficacy of our proposed methodology.

\section{Impact of channel-wise residual features}
In this part, we investigate the effect of residual features with low intensity. 
Specifically, we conduct experiments by selectively reintroducing channels from residual features that have lower average values. 
We report the results of eliminating the top $\beta$ high-value channels and the effect of normalizing $X_{\textup{res}}$ in \cref{tab:high-value}. 
The best performance is achieved when $\beta \ge 70\%$. 
Additionally, normalizing $X_{\textup{res}}$ significantly reduces its scale, resulting in performance comparable to $\beta \ge 70\%$. 
These findings support our hypothesis that high-level supervision in CLIP emphasizes global feature direction in the residual latent space, which introduces noise into the residual features.
For simplicity, we eliminate all channels in $X_{\textup{res}}$.

\begin{table}[t]
  \caption{Average performance (mIoU) over all 8 datasets.
  }
  \tabcolsep7pt
  \label{tab:high-value}
  \centering
  \begin{tabular}{l|ccccccc|c}
    \toprule
    $\beta$ (\%) & 0 & 5 & 10 & 30 & 50 & 70 & 100 & Norm \\
    \midrule
    Avg. & 22.1 & 30.2 & 33.5 & 37.4 & 38.0 & 38.1 & 38.1 & 38.1 \\
  \bottomrule
  \end{tabular}
\end{table}

\begin{table}[t]
  \caption{Average performance (mIoU) over 5 datasets without background class based on ViT-\colorbox{lightpink}{Base} and \colorbox{lightcyan}{Large} architectures. 
  }
  \tabcolsep7pt
  \renewcommand{\arraystretch}{1.0}
  \label{tab:plug-in}
  \centering
  \begin{tabular}{c|c|c|c|c|c|l}
    \toprule
    & VOC20 & Context59 & Stuff & Cityscape & ADE20K & Avg. \\
    \midrule
    \rowcolor{lightpink} CLIP \cite{radford2021learning} & 41.8 & 9.2 & 4.4 & 5.5 & 2.1 & 12.6 \\
    \rowcolor{lightpink} +ClearCLIP & 80.9 & 35.9 & 23.9 & 30.0 & 16.7 & 37.5 \textcolor{red}{\scriptsize+24.9}\\
    \midrule
    \rowcolor{lightpink} BLIP \cite{li2022blip} & 37.3 & 7.8 & 5.4 & 4.3 & 2.0 & 11.4 \\
    \rowcolor{lightpink} +ClearCLIP & 73.5 & 31.4 & 21.3 & 23.8 & 13.5 & 32.7 \textcolor{red}{\scriptsize+21.3} \\
    \midrule
    \rowcolor{lightpink} OpenCLIP \cite{cherti2023reproducible} & 47.2 & 9.0 & 5.0 & 5.1 & 2.9 & 13.8 \\
    \rowcolor{lightpink} +ClearCLIP & 81.4 & 34.1 & 23.1 & 31.8 & 18.9 & 37.9 \textcolor{red}{\scriptsize+24.1}\\
    \midrule
    \rowcolor{lightpink} MetaCLIP \cite{xu2023demystifying} & 35.4 & 8.1 & 4.3 & 5.0 & 2.2 & 11.0 \\
    \rowcolor{lightpink} +ClearCLIP & 78.3 & 34.8 & 23.5 & 27.9 & 17.4 & 36.4 \textcolor{red}{\scriptsize+25.4}\\
    \midrule
    \rowcolor{lightpink} MaskCLIP \cite{zhou2022extract} & 74.9 & 26.4 & 16.4 & 12.6 & 9.8 & 28.0 \\
    \rowcolor{lightpink} +ClearCLIP & 61.4 & 28.3 & 18.4 & 24.7 & 13.6 & 29.5 \textcolor{red}{\scriptsize+1.8} \\
    \midrule
    \rowcolor{lightpink} SCLIP \cite{wang2023sclip} & 78.2 & 33.0 & 21.1 & 29.1 & 14.6 & 35.2 \\
    \rowcolor{lightpink} +ClearCLIP & 77.9 & 35.6 & 23.6 & 31.0 & 17.0 & 37.9 \textcolor{red}{\scriptsize+1.6}\\
    \midrule
    \rowcolor{lightpink} GEM \cite{bousselham2023grounding} & 79.9 & 35.9 & 23.7 & 30.8 & 15.7 & 37.2 \\
    \rowcolor{lightpink} +ClearCLIP & 80.2 & 36.5 & 24.4 & 30.5 & 17.4 & 37.8 \textcolor{red}{\scriptsize+0.6}\\
    \midrule
    \midrule
    \rowcolor{lightcyan} CLIP \cite{radford2021learning} & 15.8 & 4.5 & 2.4 & 2.9 & 1.2 & 5.4 \\
    \rowcolor{lightcyan} +ClearCLIP & 80.0 & 29.6 & 19.9 & 27.9 & 15.0 & 34.5 \textcolor{red}{\scriptsize+29.1}\\
    \midrule
    \rowcolor{lightcyan} BLIP \cite{li2022blip} & 22.5 & 5.8 & 2.4 & 3.8 & 1.5 & 7.2 \\
    \rowcolor{lightcyan} +ClearCLIP & 67.5 & 16.8 & 11.5 & 9.3 & 7.1 & 22.4 \textcolor{red}{\scriptsize+15.2} \\
    \midrule
    \rowcolor{lightcyan} OpenCLIP \cite{cherti2023reproducible} & 39.7 & 7.0 & 4.1 & 3.9 & 2.3 & 11.4 \\
    \rowcolor{lightcyan} +ClearCLIP & 65.3 & 27.9 & 19.5 & 26.4 & 16.0 & 31.0 \textcolor{red}{\scriptsize+19.6}\\
    \midrule
    \rowcolor{lightcyan} MetaCLIP \cite{xu2023demystifying} & 22.7 & 6.2 & 3.6 & 5.1 & 2.2 & 8.0 \\
    \rowcolor{lightcyan} +ClearCLIP & 78.2 & 30.3 & 20.5 & 25.6 & 16.4 & 34.2 \textcolor{red}{\scriptsize+26.2} \\
    \midrule
    \rowcolor{lightcyan} MaskCLIP \cite{zhou2022extract} & 30.1 & 12.6 & 8.9 & 10.1 & 6.9 & 13.7 \\
    \rowcolor{lightcyan} +ClearCLIP &  65.1 & 26.5 & 17.6 & 21.2 & 15.1 & 29.1 \textcolor{red}{\scriptsize+11.1}\\
    \midrule
    \rowcolor{lightcyan} SCLIP \cite{wang2023sclip} & 60.3 & 20.5 & 13.1 & 17.0 & 7.1 & 23.6 \\
    \rowcolor{lightcyan} +ClearCLIP & 79.2 & 30.6 & 20.5 & 27.8 & 15.6 & 34.7 \textcolor{red}{\scriptsize+15.4}\\
    \midrule
    \rowcolor{lightcyan} GEM \cite{bousselham2023grounding} & 80.3 & 26.4 & 17.6 & 22.6 & 11.6 & 31.7 \\
    \rowcolor{lightcyan} +ClearCLIP & 79.7 & 29.9 & 19.4 & 25.9 & 14.2 & 33.8 \textcolor{red}{\scriptsize+2.1}\\
  \bottomrule
  \end{tabular}
\end{table}

\section{Integration across models}
Our solution serves as a free lunch applicable to various architectures and segmentation models with \textit{\textbf{just 2-3 lines of code modification}}. 
Specifically, for MaskCLIP and SCLIP, we achieve this by eliminating the residual connection and Feed-Forward Network (FFN) of the last self-attention layer. 
For GEM, we utilize the attention output from the final layer as the final representation. 
Importantly, we preserve the original attention mechanisms of these methods.
For baseline model, \ie, CLIP, BLIP, OpenCLIP, and MetaCLIP, we enhance them by incorporating our complete solution.
The performance of different models on five datasets is summarized in \cref{tab:plug-in}. 
The results demonstrate that our solution consistently enhances the performance of existing models in open-vocabulary semantic segmentation tasks, showcasing its exceptional generalizability.

\begin{figure}[t]
  \centering
  \begin{subfigure}{0.99\linewidth}
    \includegraphics[width=1.0\linewidth]{ 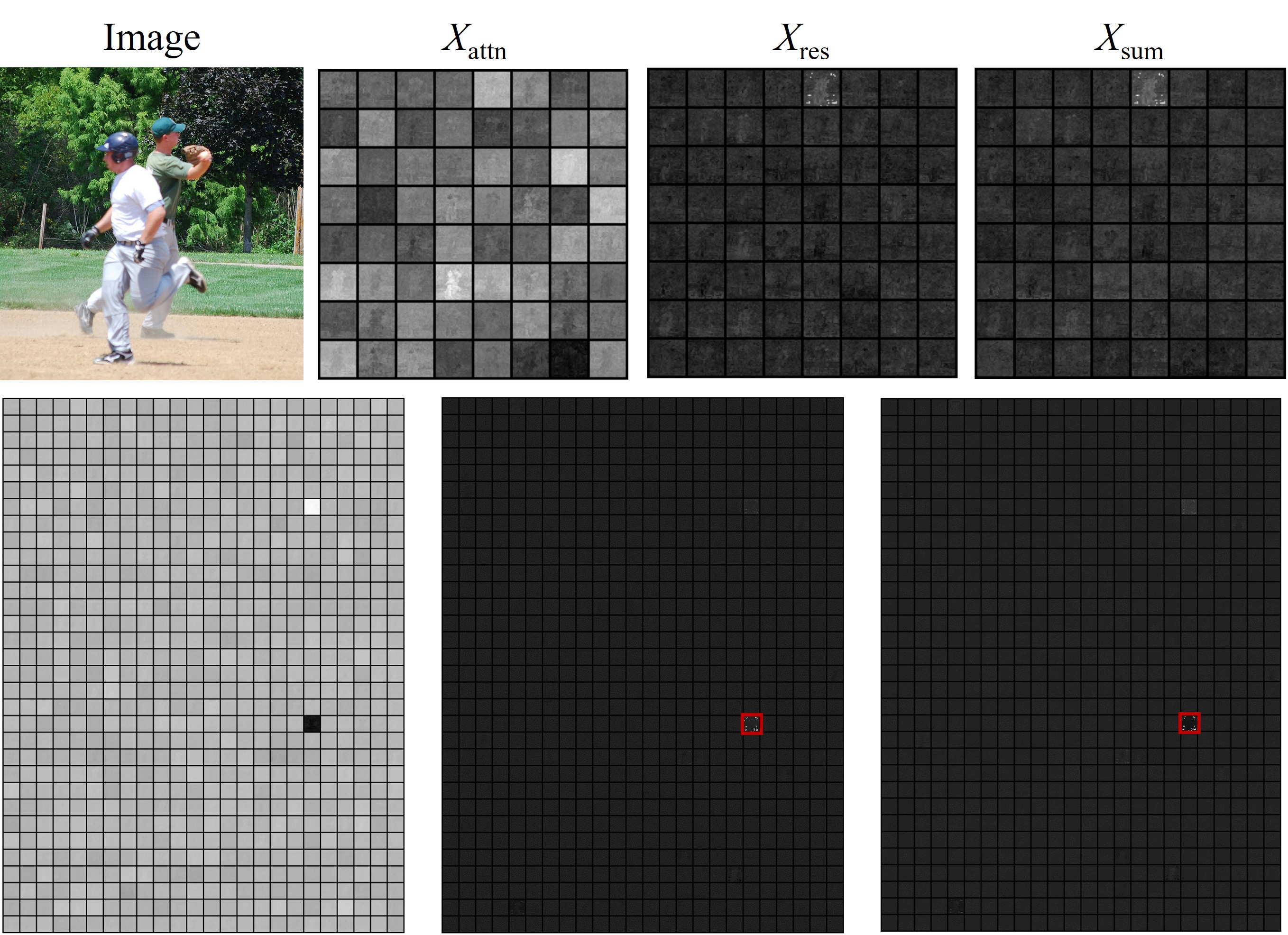}
    \label{fig:feature_maps1}
  \end{subfigure}
  \begin{subfigure}{0.99\linewidth}
  \includegraphics[width=1.0\linewidth]{ 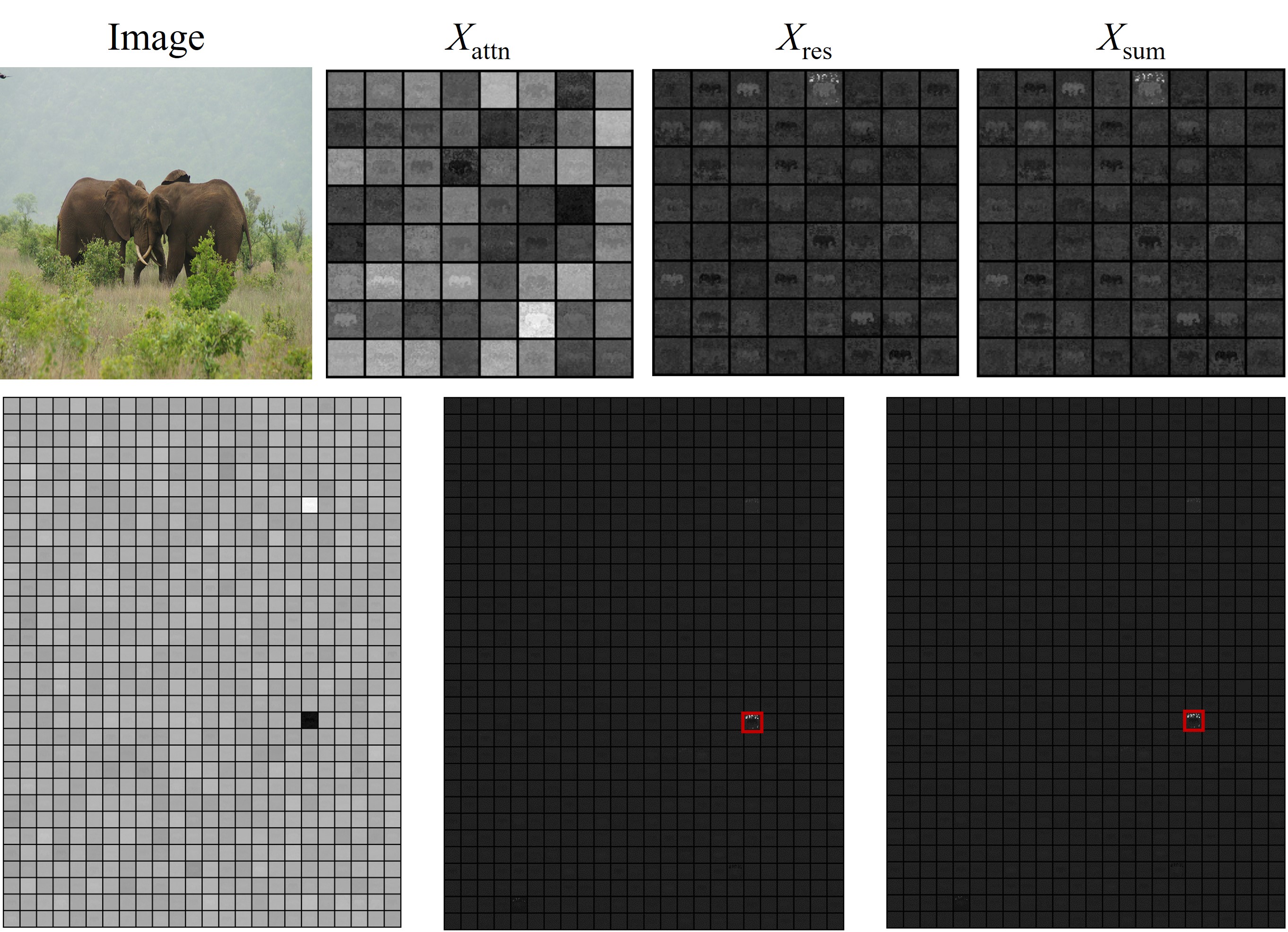}
    \label{fig:feature_maps2}
  \end{subfigure}
  \caption{Visualization of feature maps with CLIP for two randomly selected examples from the COCOStuff dataset. The first row shows the first 64 feature maps of each type, while the second row displays all 768 feature maps of each type.}
  \label{fig:feauremaps}
\end{figure}
\section{Visualization of feature maps}
\label{sec:feature maps}

To intuitively demonstrate how the residual connections affect the performance, we visualize the feature maps of $X_{\textup{res}}$, $X_{\textup{attn}}$, and $X_{\textup{sum}}$ for two randomly selected samples in \cref{fig:feauremaps}.
It is obvious that the $X_{\textup{res}}$ feature maps associated with the residual connections are characterized by peak values in one channel (highlighted in a red box), significantly surpassing the other channels. And $X_{\textup{sum}}$ is similar to $X_{\textup{res}}$, indicating the big influence of $X_{\textup{res}}$ to the final feature. Conversely, the feature maps in $X_{\textup{attn}}$ demonstrate a more uniform distribution across channels. Given that the segmentation map is derived from the cosine similarity of feature vectors at each spatial location, such a disparity implies that the features in $X_{\textup{sum}}$ and $X_{\textup{res}}$ are less discernible compared to those in $X_{\textup{attn}}$, thereby introducing noise into the segmentation results. This observation supports our proposal that \emph{the high-level supervision in CLIP emphasizes the global feature direction in the residual latent space, making local feature vectors less distinguishable and leading to noise in residual features.}

\section{Additional qualitative examples}
\label{sec:Qualitative}

\begin{figure}[t]
  \centering
  \begin{subfigure}{0.9\linewidth}
    \includegraphics[width=1.0\linewidth]{ 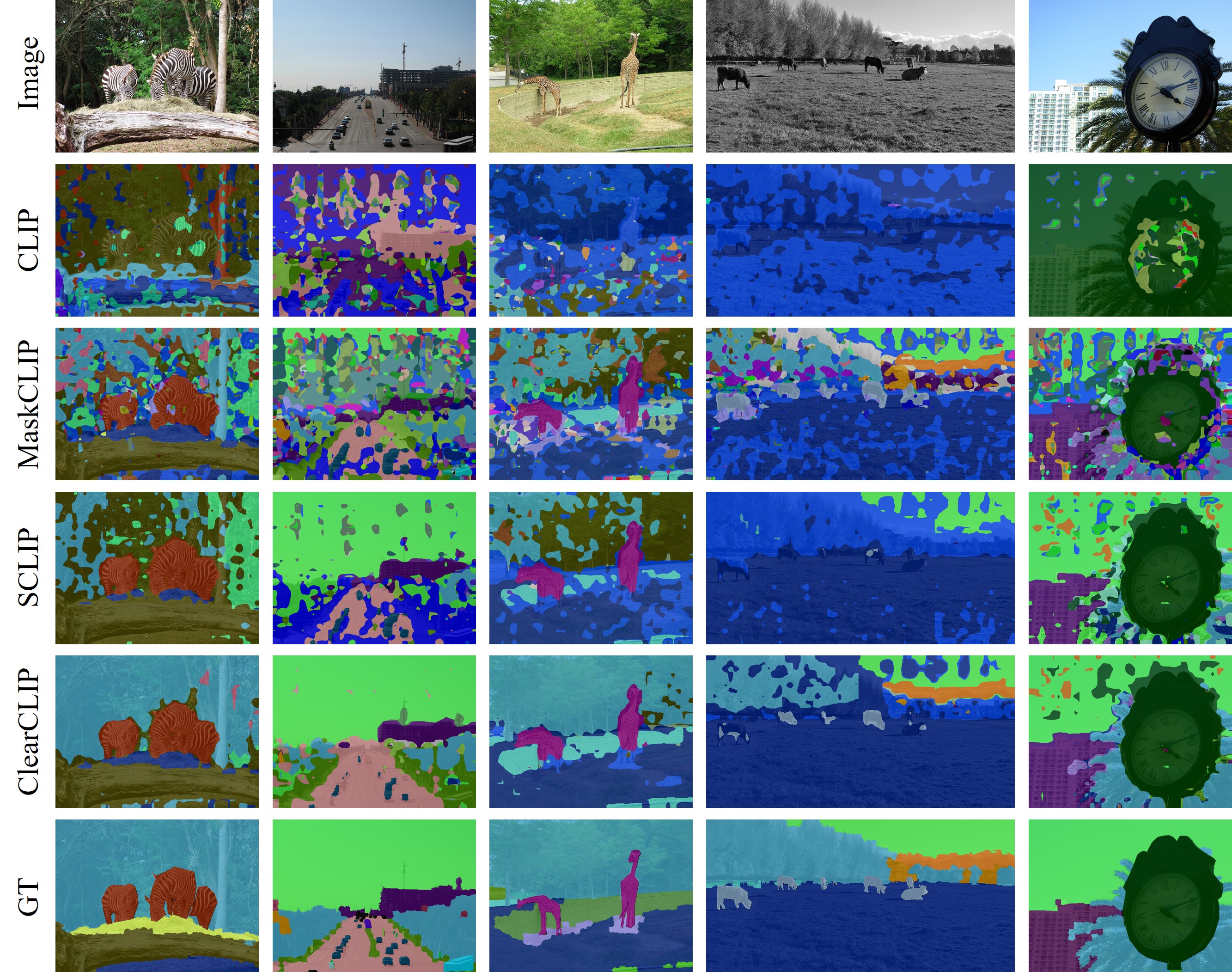}
    \caption{COCOStuff}
    \label{fig:vis_cocostuff}
  \end{subfigure}
  \begin{subfigure}{0.8\linewidth}
  \includegraphics[width=1.0\linewidth]{ 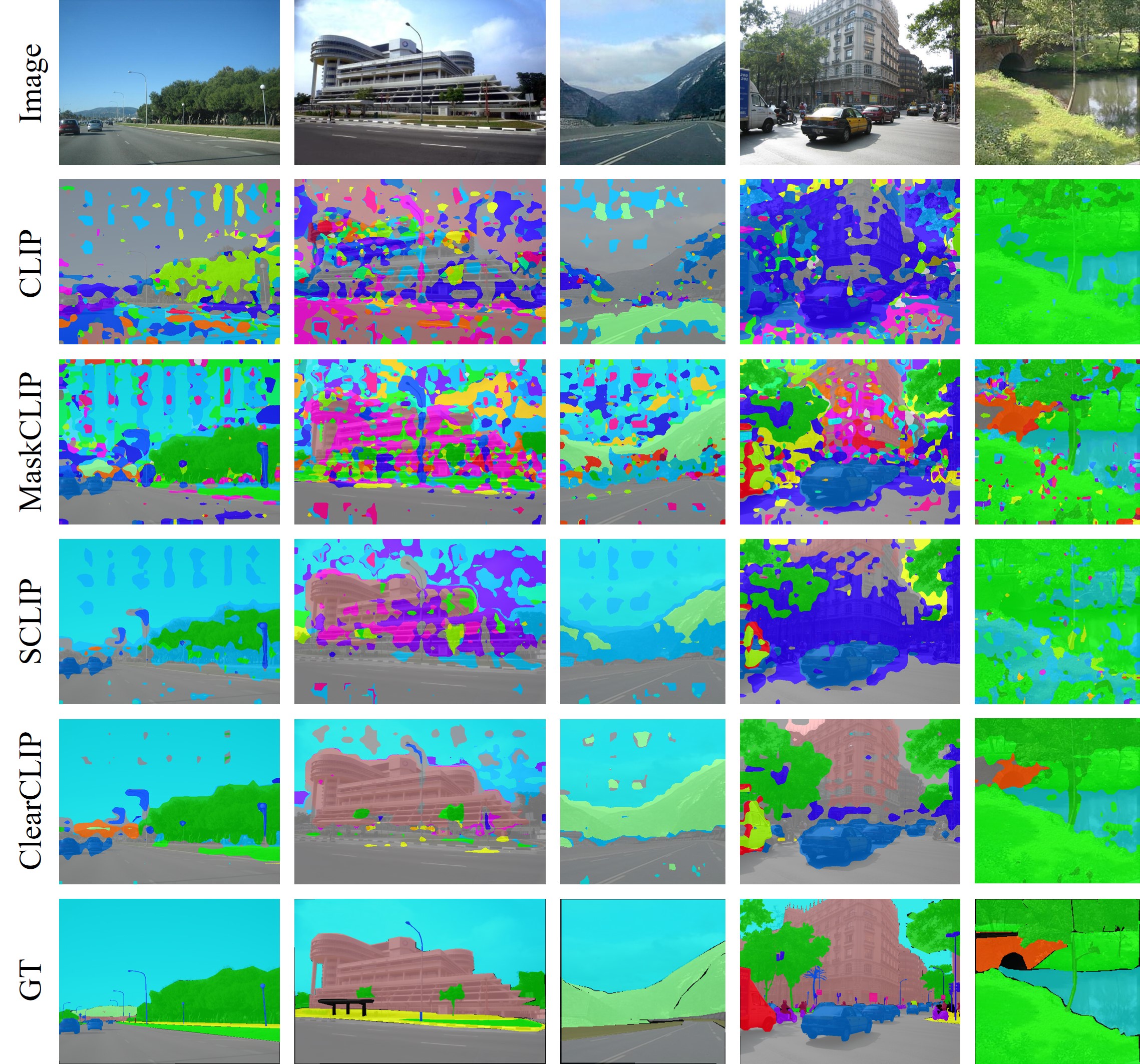}
  \caption{ADE20K}
    \label{fig:vis_ade}
  \end{subfigure}
  \caption{Qualitative comparison between different open-vocabulary segmentation methods on (a) COCOStuff and (b) ADE20K datasets.}
  \label{fig:vis_cocostuff_ade}
\end{figure}

\begin{figure}[t]
  \centering
  \includegraphics[width=1.0\linewidth]{ 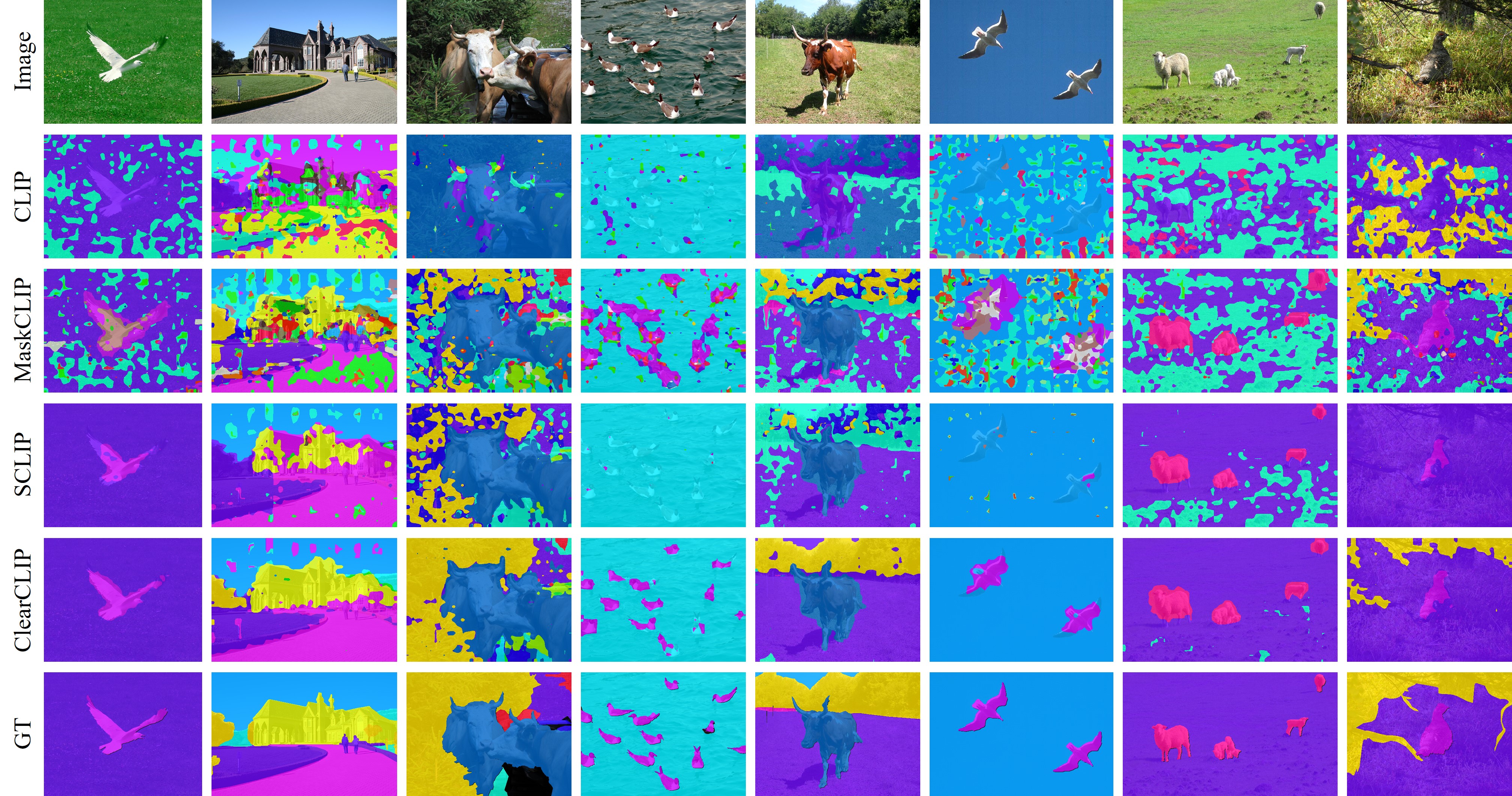}
  \caption{Qualitative comparison between different open-vocabulary segmentation methods on the Pascal Context59 dataset.}
  \label{fig:vis_context59}
\end{figure}

In this part, we present more qualitative results comparison between ClearCLIP and state-of-the-art methods. \cref{fig:vis_cocostuff_ade,fig:vis_context59} show the results from COCOStuff, ADE20K and Pascal Context59 datasets respectively. Similar to the findings in the main text, the results of ClearCLIP exhibit much less noise than other methods, further underscoring the superiority of our method.

\end{document}